%% file: main.tex
\pdfoutput=1
\documentclass[10pt,twocolumn,letterpaper]{article}

\usepackage[pagenumbers]{cvpr} 

\usepackage{graphicx}
\usepackage{amsmath}
\usepackage{amssymb}
\usepackage{booktabs}
\usepackage{multirow}
\usepackage{arydshln}
\usepackage{caption}

\usepackage{hyperref}
\hypersetup{pagebackref,breaklinks,colorlinks}

\usepackage[capitalize]{cleveref}
\crefname{section}{Sec.}{Secs.}
\Crefname{section}{Section}{Sections}
\Crefname{table}{Table}{Tables}
\crefname{table}{Tab.}{Tabs.}

\newcommand{\SF}[1]{{\color{magenta}SF: #1  }}
\newcommand{\KK}[1]{{\color{blue}Karsten: #1}}

\def\cvprPaperID{4797} 
\def\confName{CVPR}
\def\confYear{2022}

\input{preamble}
\begin{document}
\input{sec/0_metadata}

\input{fig/teaser}

\input{sec/0_abstract}
\input{sec/1_introduction}

\input{sec/2_related}

\input{sec/3_method}

\input{sec/3a_annotation}

\input{sec/4_results}
\input{sec/5_conclusions}
\input{sec/X_supplementary}

{
    \small
    \bibliographystyle{ieee_fullname}
    \bibliography{macros,main}
}


\end{document}

%% file: preamble.tex
\usepackage{overpic}
\usepackage{enumitem} 
\usepackage{overpic} 
\usepackage{color}
\usepackage{xcolor}

\definecolor{turquoise}{cmyk}{0.65,0,0.1,0.3}
\definecolor{purple}{rgb}{0.65,0,0.65}
\definecolor{dark_green}{rgb}{0, 0.5, 0}
\definecolor{orange}{rgb}{0.8, 0.6, 0.2}
\definecolor{red}{rgb}{0.8, 0.2, 0.2}
\definecolor{darkred}{rgb}{0.6, 0.1, 0.05}
\definecolor{blueish}{rgb}{0.0, 0.3, .6}
\definecolor{light_gray}{rgb}{0.7, 0.7, .7}
\definecolor{pink}{rgb}{1, 0, 1}
\definecolor{greyblue}{rgb}{0.25, 0.25, 1}

\definecolor{Highlight}{HTML}{39b54a}  

\usepackage{blindtext}

\renewcommand{\paragraph}[1]{\vspace{1em}\noindent\textbf{#1}.}

%% file: sec/0_metadata.tex

\title{\vspace{-6mm}BigDatasetGAN: Synthesizing ImageNet with Pixel-wise Annotations \vspace{-5mm}}

\author{\quad\quad Daiqing Li $^{1}$ \quad\quad Huan Ling$^{1,2,3}$   \quad\quad Seung Wook Kim$^{1,2,3}$ 
\and
  \quad Karsten Kreis$^{1}$   \quad Adela Barriuso \quad Sanja Fidler$^{1,2,3}$ \quad Antonio Torralba$^{4}$\\
  \small{\textsuperscript{1}NVIDIA \quad \textsuperscript{2}University of Toronto \quad \textsuperscript{3}Vector Institute \quad \textsuperscript{4}MIT \vspace{1pt}}\\
  \; \quad \texttt{\scriptsize \{daiqingl,huling,seungwookk,kkreis,sfidler\}@nvidia.com,} \texttt{\scriptsize   torralba@mit.edu}\\
}



%% file: fig/teaser.tex
\twocolumn[{
\renewcommand\twocolumn[1][]{#1}%
\maketitle
\begin{center}
\vspace{-9mm}
    \centering
  	\captionsetup{type=figure}
	\includegraphics[width=1\linewidth,trim=30 0 18 126,clip]{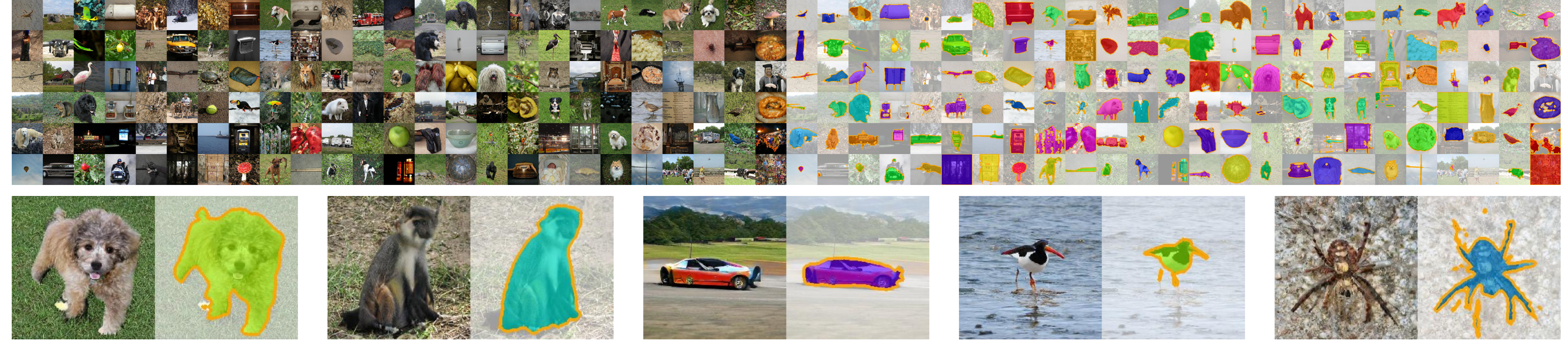}\\[-4mm]
	\captionof{figure}{Our synthesized pixel-wise labeled ImageNet dataset. We sample both images and masks for each of the 1k ImageNet classes.}
	\label{fig:teaser}

\end{center}
}]

%% file: sec/0_abstract.tex
\begin{abstract}
Annotating images with pixel-wise labels is a time-consuming and costly process. Recently, DatasetGAN~\cite{zhang2021datasetgan} showcased a promising alternative --  to \textbf{synthesize} a large labeled dataset via a generative adversarial network (GAN) by exploiting a small set of manually labeled, GAN-generated images. 
Here, we scale DatasetGAN to ImageNet scale of class diversity. We 
take image samples from the class-conditional generative model BigGAN~\cite{brock2019large} trained on ImageNet, and manually annotate only 5 images per class, for all 1k classes. 
By training an effective feature segmentation architecture on top of BigGAN, we turn BigGAN into a labeled dataset generator. We further show that VQGAN~\cite{esser2020taming} can similarly serve as a dataset generator, leveraging the already annotated data. 
We create a new ImageNet benchmark  
by labeling an additional set of real images and evaluate segmentation performance in a variety of settings. Through an extensive ablation study, we show big gains in leveraging a large generated dataset to train different supervised and self-supervised backbone models on pixel-wise tasks. 
Furthermore, we demonstrate that using our synthesized datasets for pre-training leads to improvements over standard ImageNet pre-training on several downstream datasets, such as PASCAL-VOC, MS-COCO, Cityscapes  
and chest X-ray, as well as tasks (detection, segmentation). Our benchmark will be made public and maintain a leaderboard for this challenging task. Project Page: \small{\url{https://nv-tlabs.github.io/big-datasetgan/}}

\end{abstract}

%% file: sec/1_introduction.tex
\vspace{-4mm}
\section{Introduction}
\label{sec:intro}
\vspace{-1mm}
%


The ImageNet dataset~\cite{russakovsky2015imagenet} has served as a cornerstone of modern computer vision
and deep learning.
It is used as a testbed for innovating in the domain of large-scale classification, and has enabled incredible  advancements over the years~\cite{alexnet,vggnet,He2015,dosovitskiy2020vit}. Importantly, it has also been commonly used for pre-training backbone models, with either supervised or recently  self-supervised pre-training, leading to almost guaranteed performance gains on a plethora of downstream datasets and tasks~\cite{rcnn14,zeiler13}. 
ImageNet contains a million images with 1000-way classification labels. This huge class diversity is what makes pre-trained networks generalize well to a variety of downstream applications.

\input{fig/bigdatasetgan} 

In this paper, 
we aim to enhance ImageNet with pixel-wise labels, to enable large-scale multi-class segmentation challenges and offer opportunities for new pre-training strategies for dense downstream prediction tasks. 
However, instead of manually labeling masks for 1M images, which is time-consuming and costly, we instead synthesize high quality labeled data at a fraction of the cost.

We build on top of DatasetGAN~\cite{zhang2021datasetgan}, which introduced a simple idea: to manually annotate a very small set of GAN-generated images with pixel-wise labels, and add a shallow segmentation branch on top of the GAN's feature maps which is trained on this small dataset. It was shown that the generator's feature maps are incredibly powerful and semantically meaningful, and allow the segmentation branch to produce very accurate labels for new random samples from the GAN. This means that the GAN is successfully re-purposed into a dataset generator, producing samples in the form of images and their pixel-wise labels. The authors showed that synthesizing a large dataset and using it to train downstream segmentation networks leads to extremely high performance at only a fraction of the labeling cost. However, DatasetGAN utilized StyleGAN~\cite{karras2019style} as the generative model, which is limited to single class modeling. 

In this paper, we scale DatasetGAN to the ImageNet scale, propose a novel pixel-wise ImageNet benchmark, and perform an extensive analysis of several existing methods on this benchmark. Specifically, we adopt the class-conditional generative model BigGAN~\cite{brock2019large}, which has been shown to produce high quality image samples for the 1k ImageNet classes. By manually labeling a handful of sampled images per class using a single expert annotator to ensure consistency and accuracy, we are able to synthesize a high quality labeled synthetic dataset. We further show that VQGAN~\cite{esser2020taming} can also serve as a dataset generator without the need for an additional annotation effort. 
We call a dataset generator for ImageNet obtained like that \textit{BigDatasetGAN}.
We show the benefits of leveraging synthesized datasets for a plethora of downstream dense prediction tasks and datasets. 
We demonstrate significant performance gains on PASCAL-VOC, MS-COCO, Cityscapes and chest X-ray for tasks such as object detection and instance segmentation, leveraging several different backbone models.
Furthermore, we annotate a held-out subset of real ImageNet with pixel-wise labels and introduce a new semantic segmentation benchmark. We compare several supervised and self-supervised methods and show that when these utilize our synthetic datasets their performance is significantly improved. Our benchmark will be hosted online, maintaining a leaderboard for a suite of segmentation challenges.

%% file: fig/bigdatasetgan.tex
\begin{figure*}[t!]
\vspace{-0.5mm}
\begin{center}
\includegraphics[width=1\linewidth]{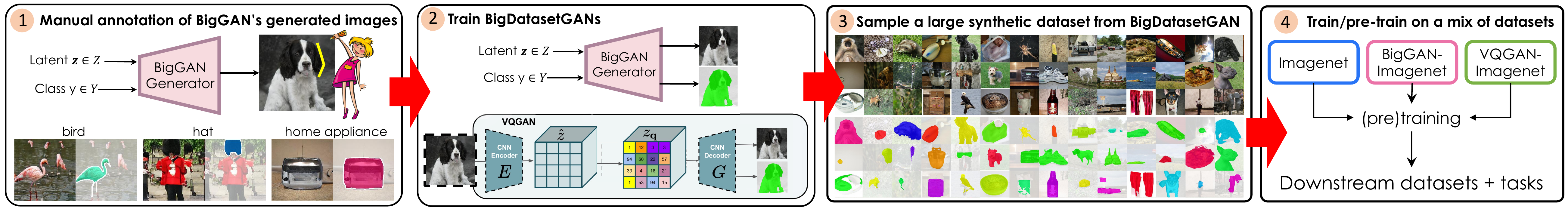}
\end{center}
\vspace{-5.5mm}
\caption{
BigDatasetGAN overview: \textit{(1)} We sample a few images per class from BigGAN and manually annotate them with masks. \textit{(2)} We train a \emph{feature interpreter} branch on top of BigGAN's and VQGAN's features on this data, turning these GANs into generators of labeled data. \textit{(3)} We sample large synthetic datasets from BigGAN \& VQGAN. \textit{(4)} We use these datasets for training segmentation models.
}
\vspace{-3mm}
\label{fig:bigdatasetgan}
\end{figure*}

%% file: sec/2_related.tex

\vspace{-1mm}
\section{Related Work}
\label{sec:related}
\vspace{-1mm}


\vspace{-3mm}
\paragraph{Reducing Annotation Cost}
Reducing labeling costs can be achieved in a variety of ways, such as interactive human-in-the-loop annotation~\cite{rother2004grabcut,PolygonPP2018,Man18,DELSE19,CurveGCN19}, nearest-neighbor mask transfer~\cite{liu2009nonparametric,tighe10,segtransfer11,autoimagenet}, or using weak supervision in the form of cheaper labels such as object boxes~\cite{ChenCVPR14,Kulharia20,Lan2021DiscoBoxWS},  scribbles~\cite{ScribbleSup,tang18} or very coarse masks~\cite{Steal19}. A full review is out of scope of this paper. 
Most related to our goal are the existing efforts to label ImageNet with pixel-wise annotations. In~\cite{autoimagenet}, the authors label 10 images per class, for 500 ImageNet classes, and iteratively propagate these labels to other images. At each stage, they perform segmentation transfer to the most similar images, and define various potentials in a graphical model to derive the final segmentation. 
While they show promising segmentation propagation performance, they do not showcase the usefulness of the auto-labeled dataset, as we do in our work.  Here, we also leverage modern deep learning techniques, GANs in particular, likely leading to higher quality datasets.

\vspace{-3mm}
\paragraph{Synthetic Dataset Generation} 
The labeling cost can also be alleviated by relying on 3D graphics engines that render images along with perfect annotations. 
Several synthetic datasets with rich labels have been created in this way~\cite{peng2017visda,VirtualHome2018,gaidon2016virtual,ros2016synthia,richter2016playing,carla}. 
However, these datasets typically exhibit a domain gap with respect to real-world datasets in terms of both appearance as well as content.  
A variety of methods translate the rendered images into more realistic ones using GAN-based techniques~\cite{munit} to close the appearance gap.
Recent works also propose to learn the parameters of the data generation pipeline 
to reduce the distribution gap~\cite{metasim,metasim2,fedsim20}.  Moreover, creating ImageNet-level of class and instance diversity via the graphics approach would require significant 3D content acquistion efforts. 

Generative models of images can be considered as a neural rendering alternative to graphics engines. High fidelity images  can now be generated by unconditional GANs such as StyleGAN~\cite{karras2019style,karras2020analyzing,karras2021aliasfree}, typically for a single class.  Class-conditional generative models such as BigGAN~\cite{brock2019large} and VQGAN~\cite{esser2020taming} have shown impressive modeling capabilities across multiple classes. The recent DatasetGAN~\cite{zhang2021datasetgan} and similar methods~\cite{xu2021linear} showed strong performance on pixel-wise segmentation when utilizing GANs to synthesize labeled data, requiring only a handful of GAN-generated images to be manually labeled. 
Works like~\cite{li2021semantic, galeev2020learning, tritrong2021repurposing} use an encoder to embed real images into a GAN's latent space and train a segmentation branch on top of the GAN's features to produce pixel-wise labeling of real images. These works mostly operate in a single-class regime, with an unconditional generative model like StyleGAN as backbone. In our work, we scale these ideas to the ImageNet scale, leveraging conditional generative models BigGAN and VQGAN, and provide extensive analysis in this setting. 

\vspace{-3mm}
\paragraph{Representation Learning}
Pre-training followed by finetuning is an effective paradigm in training neural networks. Supervised classification pre-training on large-scale datasets such as ImageNet affords large performance boosts when fine-tuning on domain-specific tasks~\cite{rcnn14,zeiler13}. 
Recently, self-supervised approaches such as \textit{contrastive learning} (CL)~\cite{chen2020simple, chen2020improved,wang2021dense, pinheiro2020unsupervised,xiao2021region} have become a widely used alternative to supervised pre-training. These methods do not require any labels, and in some cases lead to higher gains on dense downstream tasks than supervised pre-training~\cite{wang2021dense}. 
Our effort here is complementary to this line of work, and we analyze the effectiveness of these methods when combined with our synthesized datasets.

%% file: sec/3_method.tex
\input{fig/overview} 

\vspace{-1mm}
\section{BigDatasetGAN}

We extend DatasetGAN~\cite{zhang2021datasetgan} to synthesize ImageNet images with pixel-wise labels. In particular, we utilize the two ImageNet-pretrained conditional generative models BigGAN~\cite{brock2019large} and VQGAN~\cite{esser2020taming}, and enhance each with a segmentation branch, called a \emph{feature interpreter} (Sec.~\ref{sec:generative}). 
We choose these two models mainly as they represent widely-used and high-performance conditional generative models of images. We also wish to compare them since their architectures and training approaches are largely different: BigGAN has a fully convolutional architecture and is trained purely via standard adversarial objectives. VQGAN uses convolutional encoder and decoder networks, and also utilizes an autoregressive transformer on top of compressed and vector-quantized images in latent space. 

We draw a handful of samples from BigGAN and manually label them with pixel-wise annotations, a procedure which we describe in Sec.~\ref{sec:labeling}. We choose to label BigGAN's samples instead of VQGAN's for a pragmatic reason: We find that VQGAN is able to embed real images (and BigGAN's samples) with excellent reconstruction fidelity, and thus enables us to leverage annotated BigGAN samples. 
In contrast, no satisfactory encoders for BigGAN exist to date.
Using the annotated BigGAN images, we then train feature interpreter branches to predict pixel-wise segmentation labels for both BigGAN and VQGAN, following a similar approach as the original DatasetGAN~\cite{zhang2021datasetgan}.
Finally, we synthesize two large datasets by drawing labeled samples from both BigGAN and VQGAN, using filtering steps to ensure high quality of the final datasets, as described in Sec.~\ref{sec:synth_dataset}.

\subsection{Generative Models of Datasets}
\label{sec:generative}

Generative models learn a distribution 
over data, e.g., images. In the GAN framework, a generative model $\mathcal{G}$ is a function that maps latent variables $\mathbf{z}$, usually drawn from a Normal distribution $\mathbf{z}\sim \mathcal{N}(\mathbf{z};\mathbf{0},\textbf{I})$, to an image $\mathbf{x}$. Conditional GANs~\cite{brock2019large} include class information $y^c$ as input to the generator $\mathcal{G}(\mathbf{z}, y^c)$.
The generator $\mathcal{G}$ is often a convolutional neural network with sub functions $g_i$ that operate at increasingly higher spatial resolutions. We can formally write
$\mathcal{G}(\mathbf{z},y^c)= g_{l-1} \circ g_{l-2}\circ\cdots\circ g_0(\mathbf{z},y^c)$, where $l$ is the number of layers. 
If we define intermediate features $\mathbf{f}_i$ as the output of $g_i$, we obtain the set of GAN features $F_\mathcal{G}=\{\mathbf{f}_0,\mathbf{f}_1,\cdots,\mathbf{f}_{l-1}\}$ from the outputs of all the intermediate layers of $\mathcal{G}$.

We wish to learn a \emph{feature interpreter} function $\mathcal{S}(F_\mathcal{G}, y^c) \rightarrow \mathbf{y}^d$ that takes a generator's features $F_\mathcal{G}$ and a class label $y^c$ as input and outputs the set of all pixel-wise labels $\mathbf{y}^d$ 
for that class. 
$\mathcal{G}$ and $\mathcal{S}$ can be used together to generate a densely annotated dataset. 
We next discuss two different architectures, BigGAN and VQGAN, and introduce our feature interpreters' architectures on top of them. 

\input{fig/ssl}

\input{fig/dataset}
\input{fig/mean-shape}
\input{tab/dataset-analysis}

\vspace{-3mm}
\paragraph{BigGAN} BigGAN~\cite{brock2019large} adopts a convolutional architecture, illustrated in Fig.~\ref{fig:arch} on the left-hand side. Given random noise $\mathbf{z}$ and a class label $y^c$, we obtain BigGAN's generator features $F_{\textrm{BigGAN}}$ from different layers. We group features from different spatial resolutions into high-, mid- and low-level by their semantic meaning. Specifically, we group the first three ResBlocks into a high-level group with resolutions $8 \times 8$ to $32 \times 32$. We group the next three ResBlocks including one attention block together as a mid-level group with resolutions from $64 \times 64$ to $128 \times 128$. The last two ResBlocks before the image output layer are grouped into a low-level group with resolutions $256 \times 256$ to $512 \times 512$. Note that high-level features in lower layers have a very high feature dimensionality, i.e., $1536 \times 8 \times 8$.

DatasetGAN~\cite{zhang2021datasetgan} resizes all 
features into the final resolution, resulting in a large memory cost. Due to the high memory consumption, they need to randomly sample pixel-wise features when learning the MLP-based interpreter, rather than using the entire feature map. We propose to first resize features in the same group into the highest resolution within the group, and then use \texttt{1x1conv} to reduce feature dimensionality before upsampling all the features to the resolution in the next level. After \texttt{upsample}, features from lower layers are concatenated with the resized features from the current level following the same operations as described above. Features from two levels are then fused by a \texttt{mix-conv} operation which includes two \texttt{3x3conv} operations with a residual connection and a \texttt{conditional batchnorm} operation with class information, similar to \cite{brock2019large}. The same process is repeated in the low-level group, and a final \texttt{1x1conv} is used to output the segmentation logits. Compared to DatasetGAN, this design greatly reduces the memory cost and can use contextual information in the \texttt{mix-conv} operation (see Fig.~\ref{fig:arch}).

\vspace{-3mm}
\paragraph{VQGAN} One notable difference between BigGAN and VQGAN~\cite{esser2020taming} is that VQGAN incorporates an encoder that maps real images into a discrete latent space. In contrast to BigGAN, this ability allows VQGAN to exploit a labeled dataset of images that does not necessarily come from VQGAN sampling itself.  
In addition to the encoder, VQGAN has a learned class-conditional autoregressive transformer network that operates on the discrete indices of the output of the encoder. Specifically, this transformer is used to learn the discrete latent space distribution and allows a user to sample novel images from the model.
A convolutional decoder is used on top of the discrete tokens to produce an image. 
For our feature interpreter, we found it to be critical to use the features of the transformer as they contain semantic knowledge about the input image.
Specifically, we gather features from every fourth transormer layer for each spatial location $(16\times 16)$ of the encoder output.
We also use all decoder feature layers. Combining everything
yields the set of features $F_{\textrm{VQGAN}}$. We then follow the architecture design as for BigGAN
to obtain output segmentation maps.

\subsection{Synthesizing Labeled Data}
\label{sec:synth_dataset}


We sample large datasets from BigGAN and VQGAN by using several filtering steps to ensure high quality of the synthesized images and their labels. Examples are shown in Fig.~\ref{fig:alldatasets}, with the dataset analysis provided in Sec.~\ref{sec:analysis}.

\vspace{-3mm}
\paragraph{Filtering} For BigGAN, we use the \textit{Truncation Trick}\cite{brock2019large} where the noise $\mathbf{z}$ is sampled from a truncated Normal with truncation value $0.9$ to reduce noisy samples. Although lower truncation values will increase overall image fidelity and label quality as the samples are closer to the major modes of the data distribution, sample diversity is also important for downstream task performance. We further use \textit{rejection sampling} \cite{razavi2019generating}, where a pre-trained image classifier is used to rank the samples by their confidence scores, to achieve finer control over the trade-off between sample diversity and quality. We use a rejection rate of $0.9$ in our experiments. Since only 5 images on average are annotated for each class, the segmentation branch is likely to overfit.
In order to alleviate this issue, we follow \cite{zhang2021datasetgan} to train an ensemble model of 16 segmentation heads and use the Jensen-Shannon divergence as uncertainty measure to filter out the top 10\% most uncertain images~\cite{melville2005active,beluch2018ensembles,kuo2018costsensitive}. 
For VQGAN, we use top-200 filtering and nucleus sampling~\cite{holtzman2019curious} with $p=0.92$ which samples from the top-$p$ portion of the probability mass, using only the top 200 out of 16k indices.  

\vspace{-1mm}
\paragraph{Offline vs Online Sampling}
Synthesizing a static dataset incurs a one-time computational cost, and allows us to compare different segmentation methods on the same dataset. 
We also explore an online strategy, where our BigDatasetGAN is used to synthesize data online during training of a downstream segmentation model. This exposes the model to a much greater variety of data, as it never sees the same labeled examples twice during training.
In our experiments, this strategy improves task performance by 1-2\% for our ImageNet segmentation benchmark over using a static 100k-sized, sampled dataset. Compared to offline sampling, training with online sampling converges faster, as observed previously
\cite{nakkiran2021deep}. However, it
is slower, as in each training iteration
the generative model needs to be run. Therefore, we explore this approach only with the BigGAN-based model, as VQGAN sampling is particularly slow due to its autoregressive transformer component. Furthermore, we also drop expensive filtering methods like using an ensemble model in the interest of computational efficiency when performing online sampling.
We provide an analysis of the online sampling strategy for one selected method in Tab.~\ref{tab:imagenet-seg}, denoted with \emph{BigGAN-off} (offline generated dataset) and \emph{BigGAN-on} (online sampling).
Whenever not explicitly mentioned, our experiments use the offline sampling strategy. 



%% file: fig/overview.tex
\begin{figure}
\begin{center}
\includegraphics[width=.91\columnwidth]{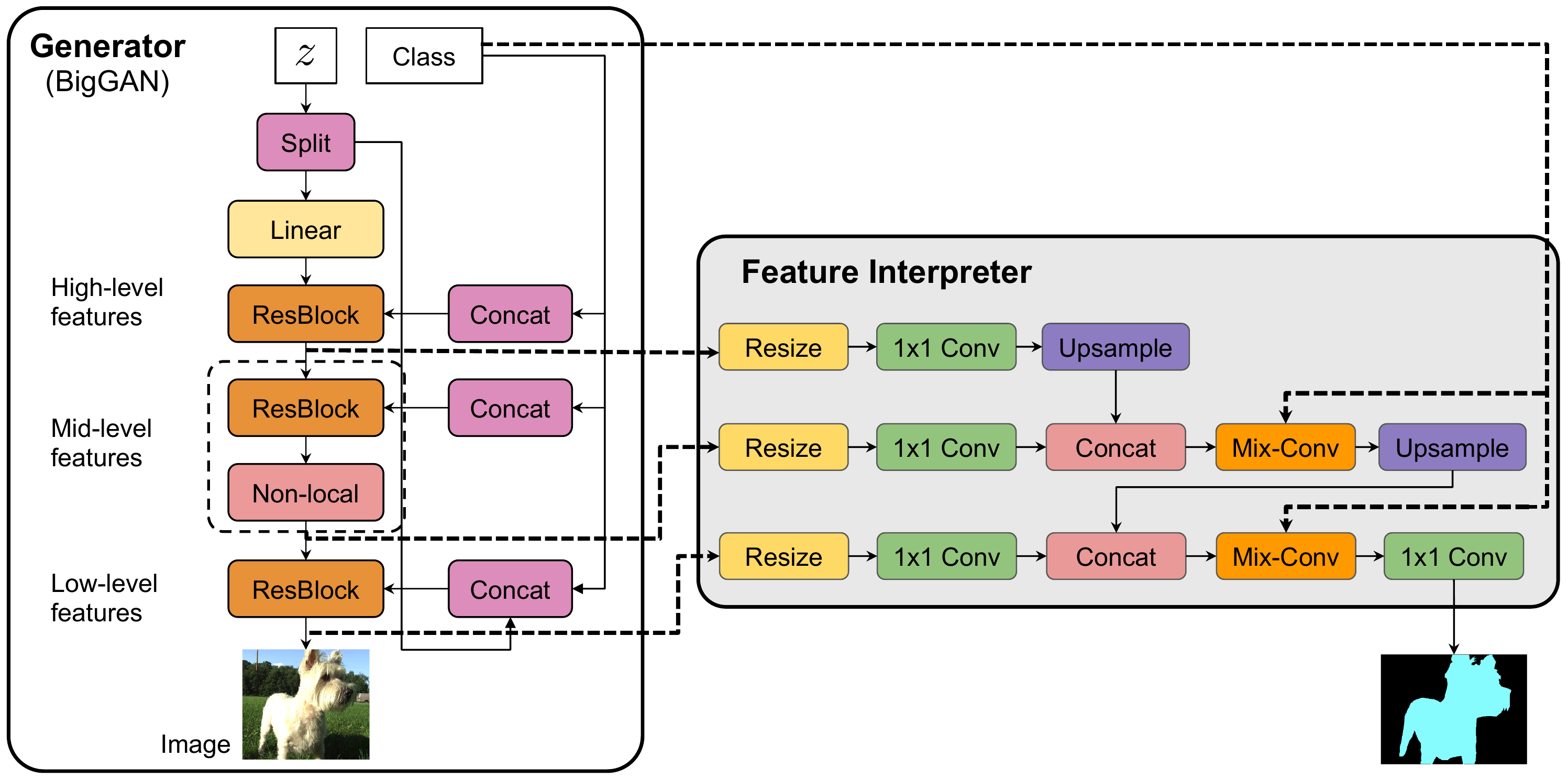}
\end{center}
\vspace{-5.5mm}
\caption{
Architecture of BigDatasetGAN based on BigGAN~\cite{brock2019large}.
}
\vspace{-3mm}
\label{fig:arch}
\end{figure}

%% file: fig/ssl.tex
\begin{figure}
\vspace{-0.5mm}
\begin{center}
\includegraphics[width=.9\columnwidth]{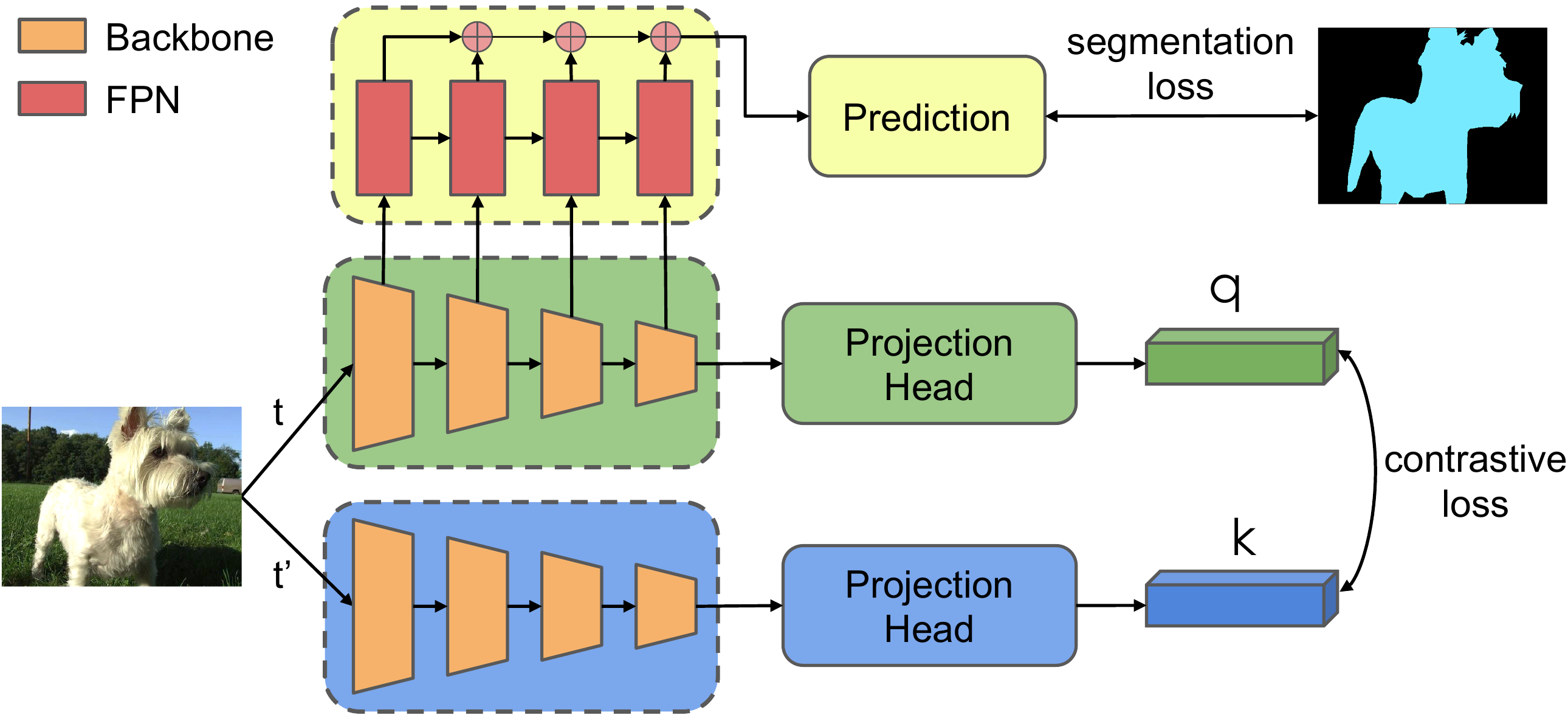}
\end{center}
\vspace{-6mm}
\caption{
Simple architecture for adding a supervised segmentation branch to self-supervised representation learners.
}
\label{fig:sslarch}
\vspace{-4.5mm}
\end{figure}

%% file: fig/dataset.tex
\begin{figure*}[t!]
\begin{center}
\includegraphics[width=\linewidth,trim=90 0 90 0,clip]{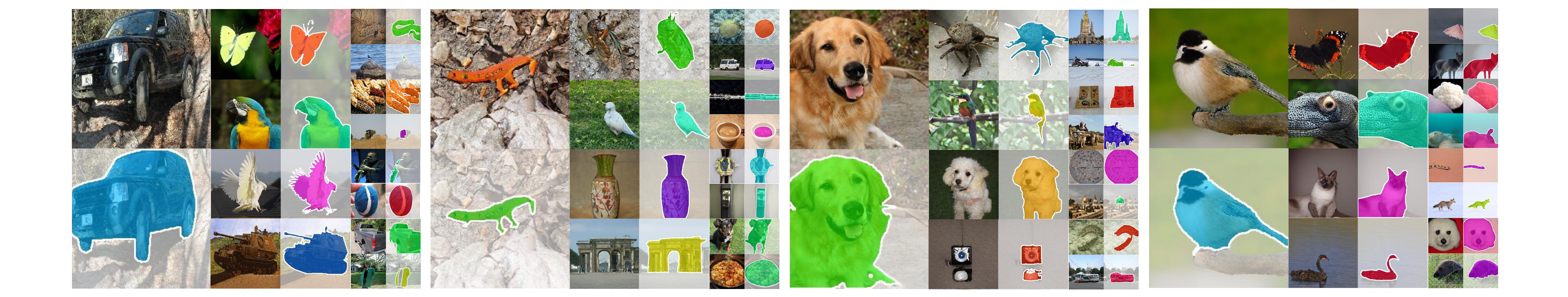}
\end{center}
\vspace{-4mm}
\begin{tabular}{p{0.24\linewidth}p{0.24\linewidth}p{0.24\linewidth}p{0.24\linewidth}}
\hspace{6.5mm}(a) Real-annotated & \hspace{1.5mm}(b) Synthetic-annotated & \hspace{2mm}(c) BigGAN-sim\hspace{1mm} & (d) VQGAN-sim
\end{tabular}
\vspace{-5mm}
\caption{
\textbf{Examples from our datasets:} \emph{Real-annotated} (real ImageNet subset labeled manually), \emph{Synthetic-annotated} (BigGAN's samples labeled manually), and synthetic \emph{BigGAN-sim}, \emph{VQGAN-sim} datasets. Notice the high quality of synthetic sampled labeled examples.
}
\label{fig:alldatasets}
\vspace{-3mm}
\end{figure*}

%% file: fig/mean-shape.tex
\begin{figure*}[t!]
\begin{center}
\includegraphics[width=1.0\linewidth]{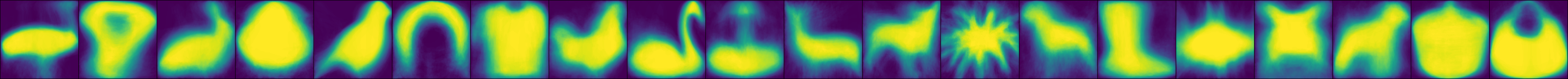}
\end{center}

\vspace{-5mm}
\caption{
\textbf{Mean shapes from our BigGAN-sim dataset.} For our 100k BigGAN-sim dataset, each class has around 100 samples. We crop the mask from the segmentation label and run k-means with 5 clusters to extract the major modes of the selected ImageNet class shapes.
}
\label{fig:meanshapes}
\vspace{-3mm}
\end{figure*}

%% file: tab/dataset-analysis.tex
\begin{table*}[t!]
\centering
\resizebox{0.9\linewidth}{!}{ 
\setlength{\tabcolsep}{8pt}
\renewcommand{\arraystretch}{1.2}
\begin{tabular}{l|c|c|c|c|c|c|c|c|c|c|c|c}

 & 
\multicolumn{5}{c|}{Dataset Statistics}   &
\multicolumn{2}{c|}{Image Quality} &
\multicolumn{2}{c|}{Label Quality} &
\multicolumn{3}{c}{Geometry Analysis}\\

\hline
Dataset & Size & \textit{IN} & \textit{MI} & \textit{BI} & \textit{MB} & \textit{FID} $\downarrow$\ & \textit{KID} $\downarrow$ & \textit{FID} $\downarrow$ & \textit{KID} $\downarrow$ & \textit{PL} & \textit{SC} & \textit{SD} \\
\hline
Real-annotated          & 8k & 1.91 & 0.315 & 0.501 & 0.581 & 0.0 & 0.0 & 0.0 & 0.0 & 3.89 & 34.4 & 7.98 \\
Synthetic-annotated     & 5k & 1.15 & 0.314 & 0.449 & 0.665 & 17.91 & 1.84  & 20.69 & 6.33 & 3.77 & 28.1 & 6.13 \\
BigGAN-sim              & 100k & 1.33 & 0.261 & 0.403 & 0.606 & 19.45 & 3.47 & 44.88 & 25.22 & 3.65 & 33.9 & 6.67 \\
VQGAN-sim               & 100k &  1.52 & 0.375 & 0.615 & 0.583 & 21.21 & 11.10 & 48.55 & 33.28 & 3.81 & 30.8 & 5.13 \\
\end{tabular}}
\vspace{-3mm}
\caption{
\textbf{Dataset analysis.} We report image \& mask statistics across our datasets (naming explained in Fig.~\ref{fig:alldatasets}). 
We compute image and label quality using FID and KID and use Real-annotated dataset as reference. \textit{IN}: instance count per image, \textit{MI}: ratio of mask area over image area, \textit{BI}: ratio of tight bounding box of the mask over image area, \textit{MB}: ratio of mask area over area of its tight bounding box, \textit{PL}: polygon length (polygon normalized to width and height of 1), \textit{SC}: shape complexity measured by the number of points in a simplified polygon, \textit{SD}: shape diversity measured by mean pair-wise Chamfer distance~\cite{fan2016point} per class and averaged across classes.
} 
\label{tab:dataset-analysis}
\end{table*}

%% file: sec/3a_annotation.tex
\vspace{-2mm}
\section{Annotating ImageNet with Pixel-wise Labels}
\label{sec:labeling}

The first important choice to consider is what data to label manually to form the training set for our feature interpreters. While the GANs ensure that we can operate in the low-label regime, labeling all 1k ImageNet classes still incurs a cost, mainly in labeling time as we utilize a single annotator. 
Since BigGAN does not have the ability to encode images and thus utilize any labeled data -- 
in contrast to VQGAN, which has an encoder --
we opt to label BigGAN's samples. Note that BigGAN's samples are mostly very realistic and diverse. 
We highlight that BigGAN produces samples 150 times faster than VQGAN, and is thus more convenient for more massive-scale experiments.

From each category, we randomly selected 10 images for annotation from both the ImageNet, and BigGAN generated images. For each of the 1k categories, a single 
annotator
segmented both the real and synthetic images. Both sets were annotated consecutively starting with the real images and then segmenting the BigGAN images to facilitate the recognition of the synthetic images. The annotator labeled 5 images per class and dataset on average, ignoring low quality images with unrecognizable objects. Out of the 1k categories, only for 3 categories the annotator could not identify any of the images generated by BigGAN.

%% file: sec/4_results.tex
\input{tab/top5}
\input{tab/imagenet-seg}

\input{fig/combined_ablations}

\vspace{-2mm}
\section{Analysis and Experiments}
\label{sec:results}


In Sec.~\ref{sec:analysis}, we provide analyses of our synthesized datasets compared to the real annotated ImageNet samples and highlight new insights into ImageNet through segmentation.
In Sec.~\ref{sec:benchmark}, we evaluate several state-of-the-art supervised and self-supervised representation learning methods by competing on our (real) pixel-wise ImageNet benchmark and ablate the use of our synthetic datasets. Finally, in Sec.~\ref{sec:transfer} we showcase the benefits of our synthesized data for pre-training of downstream models on various datasets. Further experiment details and analyses in Supp. Material.

\vspace{-1mm}
\subsection{Dataset Analysis}
\label{sec:analysis}
We first study our four different datasets in Table~\ref{tab:dataset-analysis}: \textit{(1)} Real manually annotated dataset (size 8k), which is used as benchmark testing data. \textit{(2)} Synthetic manually annotated BigGAN dataset (size 5k), which is used as benchmark training data. And two synthetic datasets \textit{(3)} BigGAN-sim (size 100k) and \textit{(4)} VQGAN-sim (size 100k) generated from BigGAN and VQGAN, respectively. We compare image and label quality in terms of distribution metrics using the real annotated dataset as reference. We also compare various label statistics and perform shape analysis on labeled polygons in terms of shape complexity and diversity.

\vspace{-3mm}
\paragraph{Synthetic vs. real dataset} Compared to the Real-annotated dataset, the Synthetic-annotated dataset shows a  distribution gap in terms of FID/KID~\cite{heusel2017fid,binkowski2018demystifying} on sampled images masked by segmentation labels (Table~\ref{tab:dataset-analysis}, also see Fig.~\ref{fig:alldatasets} and Fig.~\ref{fig:meanshapes}). We compute metrics on masked images to avoid noisy backgrounds. Among the synthetic datasets, BigGAN-sim has better image and label quality in terms of FID/KID compared to VQGAN-sim ($19.45/3.47$ vs. $21.21/11.10$). However, VQGAN-sim contains on average 1.52 instances per image (IN) compared to BigGAN-sim (1.33). This indicates that VQGAN can better model multiple objects in individual images. Interestingly, BigGAN-sim labels have a higher Shape Complexity (SC) score compared to VQGAN-sim ($33.9$ vs. $30.8$) measured by the average number of points in polygons simplified by the Douglas-Peucker algorithm~\cite{douglas1973algorithms} with threshold 0.01. The Shape Diversity (SD) metric is also slightly better for BigGAN. Note that we may be seeing some of the effects of training VQGAN on BigGAN's data. 

\subsection{ImageNet Segmentation Benchmark}
\label{sec:benchmark}

We introduce a benchmark with a suite of segmentation challenges using our Synthetic-annotated dataset (5k) as training set and evaluate on our Real-annotated held-out dataset (8k); please refer to Table \ref{tab:dataset-analysis} for details. Specifically, we evaluate performance for \textit{(1)} two individual classes (dog and bird), \textit{(2)} foreground/background (FG/BG) segmentation evaluated across all 1k classes, and \textit{(3)} multi-class semantic segmentation for various subsets of classes. 

\input{fig/benchmark_testing/benchmark_testing}

\vspace{-3mm}
\paragraph{Task setup} 
\textit{Dog} and \textit{Bird} evaluate 
binary
segmentation. For \textit{Dog}, we group 118 dog classes in ImageNet1k, resulting in 657 training images. For \textit{Bird}, we group 59 bird classes, with 366 training images. The FG/BG task evaluates
binary
segmentation accuracy over all classes, while \textit{MC-16} is multi-class segmentation on a group of 16 common objects like boat, car and chair, similar to PASCAL VOC~\cite{Everingham10}. \textit{MC-100} is also multi-class segmentation over randomly selected 100 ImageNet1k classes. Task \textit{MC-128} uses top-down grouping based on WordNet~\cite{huh2016makes}, resulting in a long-tailed class distribution, suitable for testing class-imbalanced segmentation. \textit{MC-992} is a multi-class segmentation task on all 992 ImageNet1k classes, where we filtered out 8 classes that BigGAN cannot model well. 

\vspace{-3mm}
\paragraph{Evaluation} All segmentation methods we compare are based on DeepLabv3~\cite{deeplabv32018} with Resnet-50~\cite{He2015}. The evaluation metric is mean Intersection-over-Union (mIoU). 

\vspace{-3mm}
\paragraph{Comparisons} In Table~\ref{tab:imagenet-seg}, we compare state-of-the-art self-supervised learning (SSL) methods based on either contrastive learning or knowledge distillation. We also compare performance over supervised pre-training on ImageNet. Since SSL methods do not use class information, we also include the backbone pre-trained using Supervised Contrastive Learning (SupCon)~\cite{khosla2021supervised} and MoCo-v2 jointly trained with classification (SupSelfCon)~\cite{islam2021broad}. We observe that whenever a method is trained using our large synthetic datasets (in contrast to only using the manually labeled dataset) the task performance improves. Specifically, the MoCo-v3 pre-trained backbone trained using the VQGAN-sim dataset improves mean performance by \textbf{10.2} over the 7 tasks. In \textit{MC-16}, the DenseCL~\cite{wang2021dense} pre-trained backbone improves its performance by \textbf{11.4} with our synthetic dataset.

\vspace{-3mm}
\paragraph{VQGAN-sim vs. BigGAN-sim} In our benchmark, VQGAN-sim and BigGAN-sim have the same dataset size of 100k labeled images. In terms of task performance, methods trained with VQGAN-sim achieve overall better performance. However, VQGAN is a massive model with 1.5B parameters including the transformer, while BigGAN has only 110M parameters. In terms of inference speed, since VQGAN samples autoregressively, it takes on average 15 sec. per image, whereas BigGAN's inference speed is around 0.1 sec. per image. Sampling a 2M synthetic dataset would take almost an entire year for VQGAN vs. 55 hours for BigGAN on a single GPU. As discussed, this also makes it impractical
to use the online sampling strategy with VQGAN during training of downstream task models.

\vspace{-3mm}
\paragraph{Scaling up dataset size} A useful property of  generative models is the ability to synthesize large amounts of data. In Fig.~\ref{fig:datasize}, we show that task performance increases when scaling up dataset size. A model trained with a 22k-sized synthetic dataset from BigGAN-sim outperforms the same model trained with 2k human-annotated labels. We observe another gain of 7 points when scaling dataset size from 22k to 220k. A further but less significant boost is seen at 2M.

\vspace{-3mm}
\paragraph{Scaling up model size} We also analyze the effect of the model size. In Fig.~\ref{fig:modelsize}, we compare a baseline model trained with 2k human-annotated labels with the same model trained with the 100x larger BigGAN-sim dataset. We see large performance gains at all model sizes when leveraging our large simulated dataset.

\vspace{-3mm}
\paragraph{Classification vs. segmentation} Furthermore, we study whether shapes which are difficult to segment are also hard to classify. In Table~\ref{tab:top5-analysis}, we show the Top-5 worse/best classes in terms of class-wise FG/BG segmentation. We find that classes with thin structures, like bow, as well as objects with heavy occlusions and no clear boundaries, are hard to segment but not necessarily difficult to classify.

\input{tab/mscoco}
\input{tab/voc-det}
\input{tab/cxr-seg}
\input{tab/cityscape-voc}

\vspace{-1mm}
\subsection{Transfer Learning}
\label{sec:transfer}

Here, we propose a simple architecture design to jointly train model backbones with contrastive learning and supervision from our synthetic datasets (Fig.\ref{fig:alldatasets}). We conduct experiments with SSL approaches for dense prediction tasks on MS-COCO\cite{lin2015microsoft}, PASCAL-VOC\cite{everingham2010pascal}, Cityscapes\cite{cordts2016cityscapes}, as well as chest X-ray segmentation in the medical domain.

\vspace{-3mm}
\paragraph{Pre-training with synthetic data} We build on top of state-of-the-art contrastive learning method DenseCL~\cite{wang2021dense}, which extends MoCo-v2~\cite{chen2020improved} with a dense contrastive loss. We design a shallow semantic segmentation decoder based on feature pyramid networks (FPN), similar to~\cite{kirillov2019panoptic}, to train a segmentation branch on our synthetic dataset (Fig.~\ref{fig:sslarch}). We follow the same training and augmentation protocol as DenseCL and use the same hyperparameters for pre-training. To speed up pre-training, we start from the checkpoint trained on ImageNet after 150 epochs and then jointly train with the segmentation and contrastive losses for another 50 epochs. Thus, for fair comparison, we compare with baselines pre-trained for 200 epochs. We use Resnet-50 as backbone and BigGAN-sim 100k dataset to train segmentation branch in all SSL experiments. After pre-training, only the backbone is used on downstream tasks.

\vspace{-3mm}
\paragraph{COCO object detection and instance segmentation} We use Mask R-CNN~\cite{he2017maskrcnn} with Resnet-50 FPN backbone. Only the pre-trained Resnet-50 backbone is used during fine-tuning. We follow the setup from~\cite{he2020momentum} to fine-tune all the layers on  \texttt{train'17} and evaluate on \texttt{val'17}. We use the standard $1\times$ and $2\times$ training schedules implemented in Detectron2~\cite{wu2019detectron2}. For the $1\times$ schedule we improve $AP^{bb}$ by 0.4 points and $AP^{mk}$ by 0.3 points when trained with our BigGAN-sim dataset (Table~\ref{tab:mscoco-det}). We see further performance improvements when training longer with the $2\times$ schedule, consistently improving over the baseline methods.

\vspace{-3mm}
\paragraph{PASCAL VOC object detection and semantic segmentation} For object detection, we use Faster R-CNN~\cite{fasterrcnn} with R50-C4~\cite{he2017maskrcnn} backbone implemented in Detectron2. For semantic segmentation, we follow~\cite{wang2021dense} to train a FCN~\cite{long2015fully} model on \texttt{train\_aug'12} (10582 images) for 20k iterations and evaluate on \texttt{val'12} implemented in MMSegmentation~\cite{mmseg2020}. Compared to DenseCL in Table~\ref{tab:voc-det}, our synthetic data for pre-training improves detection by 0.3 $AP_{75}$ and segmentation performance by 0.5\% on mIoU. We also find significantly faster convergence rates compared to training with contrastive learning alone (Fig.~\ref{fig:voc-seg}).

\vspace{-3mm}
\paragraph{Cityscapes instance segmentation and semantic segmentation} We use Mask R-CNN \cite{he2017maskrcnn} with Resnet-50 FPN backbone trained on standard splits \cite{cordts2016cityscapes} for 90k iterations. In Table~\ref{tab:cityscape-voc}, training with our BigGAN-sim dataset improves $AP^{mk}$ by 0.3 points in the instance segmentation task over the baseline model. However, we do not see a significant performance boost for the semantic segmentation task.

\vspace{-3mm}
\paragraph{Medical imaging with limited data} We construct the chest X-ray segmentation dataset by combining JSRT/SCR \cite{shiraishi2000development}, NLM \cite{jaeger2014two}, Shenzhen\cite{jaeger2014two} and NIH\cite{stirenko2018chest} datasets and split it into 735 train and 316 test images. To test the backbone's transferability, we freeze the weights of the backbone including the batchnorm statistics and only fine-tune the FCN head. All methods are run using Resnet-50 with batchsize 8 for 30 epochs. Since the downstream task is binary segmentation, we pre-train the network to predict FG/BG on our synthetic dataset. In Table \ref{tab:cxr-seg}, 
we see the model trained using our data achieves $67.6$ mIoU vs. $48.8$ mIoU for the baseline method with only $1\%$ training data.

%% file: tab/top5.tex
\newcommand\hhh{1.3cm}
\newcommand\www{3.4cm}
\begin{figure*}[t!]
\center
\vspace{-0mm}
\addtolength{\tabcolsep}{-4.5pt}
\begin{tabular}{ccccc}

\includegraphics[height=\hhh,width=\www, trim=0 0 0 0,clip]{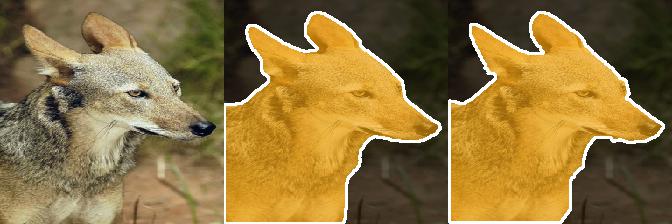} &
\includegraphics[height=\hhh,width=\www, trim=0 0 0 0,clip]{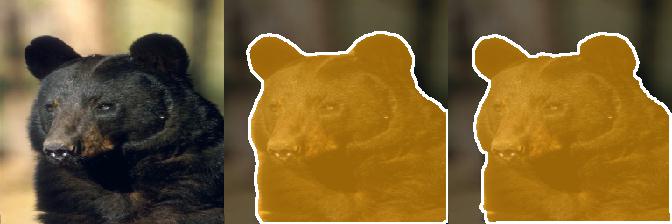} & 
\includegraphics[height=\hhh,width=\www,  trim=0 0 0 0,clip]{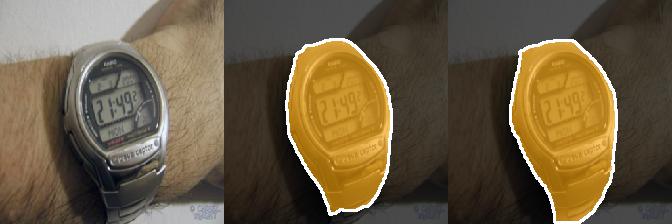} &
\includegraphics[height=\hhh,width=\www,  trim=0 0 0 0,clip]{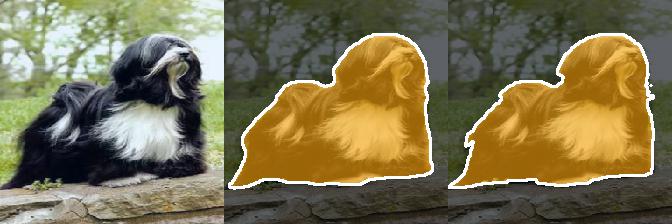} &
\includegraphics[height=\hhh,width=\www,  trim=0 0 0 0,clip]{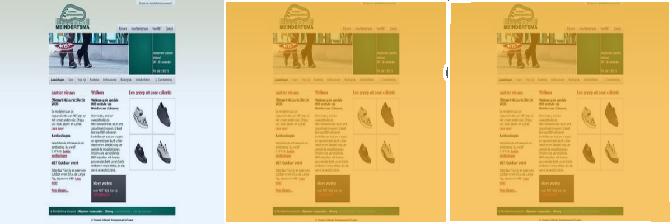} 
 \vspace{-1mm}
 \\[-1mm]
 \vspace{-0.5mm}
   {\scriptsize \texttt{red wolf}: \texttt{97.7/ 0.8}} &  {\scriptsize \texttt{black bear}: \texttt{97.5/ 1.0}} &  {\scriptsize \texttt{watch}: \texttt{95.8/ 1.0} } &  {\scriptsize \texttt{terrier}: \texttt{95.6/ 1.0}} & {\scriptsize \texttt{website}: \texttt{95.4/ 1.0}} \\
   
\includegraphics[height=\hhh,width=\www, trim=0 0 0 0,clip]{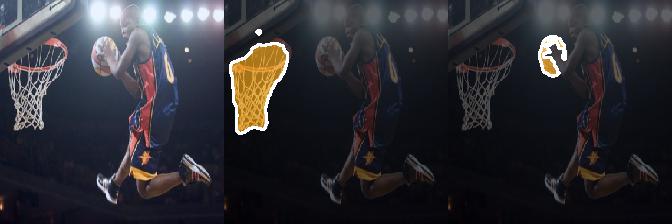} &
\includegraphics[height=\hhh,width=\www,  trim=0 0 0 0,clip]{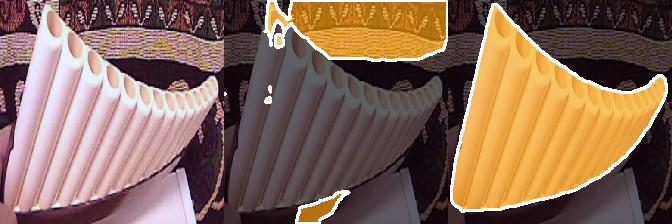}  &
\includegraphics[height=\hhh,width=\www,  trim=0 0 0 0,clip]{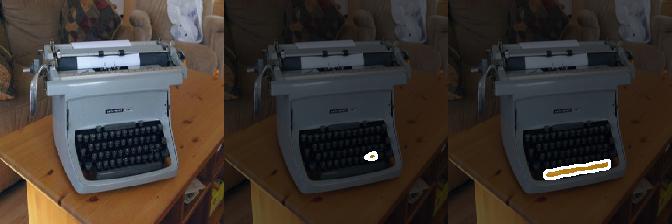} &
\includegraphics[height=\hhh,width=\www,  trim=0 0 0 0,clip]{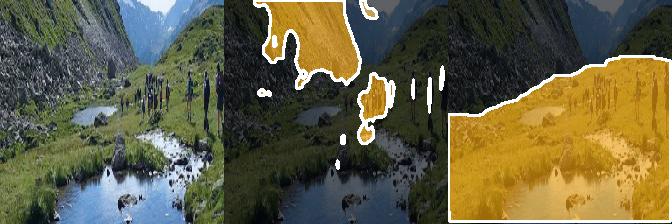} &
\includegraphics[height=\hhh,width=\www, trim=0 0 0 0,clip]{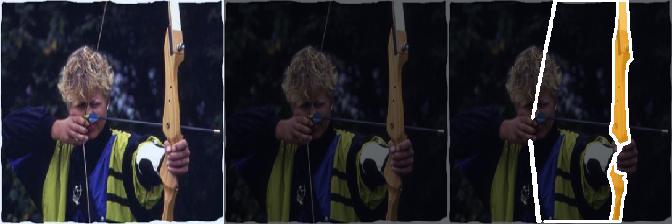}   \vspace{-1mm} \\ \vspace{-0.5mm}
  {\scriptsize \texttt{basketball}: \texttt{0.1/1.0}} &  {\scriptsize \texttt{pandean pipe}: \texttt{1.8/1.0}} &  {\scriptsize \texttt{space bar}: \texttt{2.4/0.7} } &  {\scriptsize \texttt{valley}: \texttt{2.5/0.5}} & {\scriptsize \texttt{bow}: \texttt{3.9/1.0}} \\
\vspace{-1.5mm}
\end{tabular}
\vspace{-5mm}
\caption{\footnotesize  {\bf Top-5 analysis of ImageNet benchmark.}  Text below images indicates: Class name,  \textit{FG/BG} segmentation measured in mIoU, classification accuracy of a Resnet-50 pre-trained on ImageNet. \textit{Top Row:} We visualize Top-5 best predictions of DeepLabv3 trained on BigGAN-sim dataset for the \textit{FG/BG} task, compared to ground-truth annotations (\textit{third} column).  \textit{Bottom Row:} We visualize Top-5 worst predictions. Typical failure cases include small objects or thin structures. Interestingly, classes the are hard to segment, such as \texttt{baskeball} and \texttt{bow}, are not necessarily hard to classify.
}
\label{tab:top5-analysis}
\end{figure*}

%% file: tab/imagenet-seg.tex
\begin{table}[t]
\centering
\resizebox{1.02\linewidth }{!}{ 
\setlength{\tabcolsep}{2pt}
\renewcommand{\arraystretch}{1.1}

\hspace{-2.5mm}\begin{tabular}{c|l|c|c|c|c|c|c|c||c}
& Method  & \textit{Dog} & \textit{Bird} & \textit{FG/BG} & \textit{MC-16} & \textit{MC-100} & \textit{MC-128} & \textit{MC-992} & Mean\\
\hline
\parbox[t]{3.1mm}{\multirow{6}{*}{\rotatebox[origin=c]{90}{superv. pre-train.}}} & Rand                                        & 56.4 & 35.7 & 44.7 & 13.9 & 4.0 & 3.6 & 2.3 & 22.9\\
\cdashline{2-10} 
& Sup.IN                                      & 82.6 & 79.6 &  66.3 & 58.7 & 56.1 & 28.5 & 17.8 & 55.7\\
& $\ $ +  BigGAN-off                             & 85.8 & 81.2 & 67.5 & 64.6 & 62.3 & 29.3 & 22.8 & 59.0\\
& $\ $ +  BigGAN-on                               & \textbf{87.0} & \textbf{83.2} & \textbf{69.5} & \textbf{66.1} & \textbf{62.8} & \textbf{29.5} & \textbf{24.6} & \textbf{60.4}\\
\cdashline{2-10} 
 & SupCon~\cite{khosla2021supervised}           & 83.8 & 79.0 & 66.6 & 59.2 & 55.4 & 28.6 & 18.7 & 55.9\\
& SupSelfCon~\cite{islam2021broad} & 84.4 & 81.8 & 67.6 & 63.1 & 60.0 & 28.3 & 18.9 & 57.8\\
& $\ $ +  BigGAN-sim                              & \textbf{87.0} & 83.2 & 69.5 & 66.1 & 62.8 & \textbf{32.8} & \textbf{29.7} & \textbf{61.6}\\
& $\ $ +  VQGAN-sim                               & 86.7 & \textbf{84.4} & \textbf{71.1} & \textbf{68.1} & \textbf{64.7} & 30.4 & 25.8 & \textbf{61.6}\\

\hline
\parbox[t]{3.1mm}{\multirow{10}{*}{\rotatebox[origin=c]{90}{self-superv. pre-train.}}} & SimCLR\cite{chen2020simple}                 & 77.7 & 73.7 & 66.7 & 53.8 & 44.3 & 29.3 & 16.0 & 51.6\\
& MoCo-v2\cite{chen2020improved}               & 84.6 & 82.7 & 65.6 & 51.4 & 39.1 & 18.5 & 10.2 & 50.3\\
& BYOL\cite{grill2020bootstrap}               & 78.0 & 72.9 & 68.5 & 55.4 & 45.8 & 27.7 & 16.1 & 52.1\\
& DINO\cite{caron2021emerging}                & 77.8 & 72.7 & 66.1 & 50.5 & 41.2 & 23.2 & 12.7 & 49.2\\
\cdashline{2-10} 
& DenseCL\cite{wang2021dense}         & 85.0 & 83.3 & 62.5 & 47.9 & 38.6 & 17.3 & 13.1 & 49.7\\
& $\ $ + BigGAN-sim                              & 86.5 & 84.9 & 66.4 & 58.6 & 41.0 & 17.9 & \textbf{19.9} & 53.6\\
& $\ $ + VQGAN-sim    & \textbf{86.7} & \textbf{85.9} & \textbf{67.1} & \textbf{59.3} & \textbf{41.7} & \textbf{19.3} & 16.8 & \textbf{53.8}\\
\cdashline{2-10} 
&  MoCo-v3\cite{chen2021empirical}            & 77.2 & 71.5 & 67.4 & 54.0 & 49.7 & 30.1 & 17.4 & 52.5\\
& $\ $ +  BigGAN-sim                              & 83.3 & 76.7 & 71.1 & 64.8 & 58.1 & 38.9 & 30.8 & 60.5\\
& $\ $ +  VQGAN-sim                               & \textbf{83.8} & \textbf{80.9} & \textbf{71.7} & \textbf{66.5} & \textbf{61.2} & \textbf{41.5} & \textbf{33.5} & \textbf{62.7}\\

\end{tabular}
} 
\vspace{-3mm}
\caption{
\textbf{ImageNet pixel-wise benchmark.} We compare various methods on several tasks, with supervised and self-supervised pre-training. We use Resnet-50 for all methods. We ablate the use of synthetic datasets for three methods. \emph{FG/BG} evaluates binary segmentation across all classes; \emph{MC-N} columns evaluate multi-class segmentation performance in setups with $N$ classes. Adding synthetic datasets improves performance by a large margin \emph{BigGAN-off} and \emph{BigGAN-on} compare offline \& online sampling strategy. 
} 
\label{tab:imagenet-seg}
\vspace{-4mm}
\end{table}

%% file: fig/combined_ablations.tex
\begin{figure*}[t!]
\vspace{-2mm}
\begin{minipage}{0.32\textwidth}
\begin{center}
\includegraphics[width=.99\columnwidth]{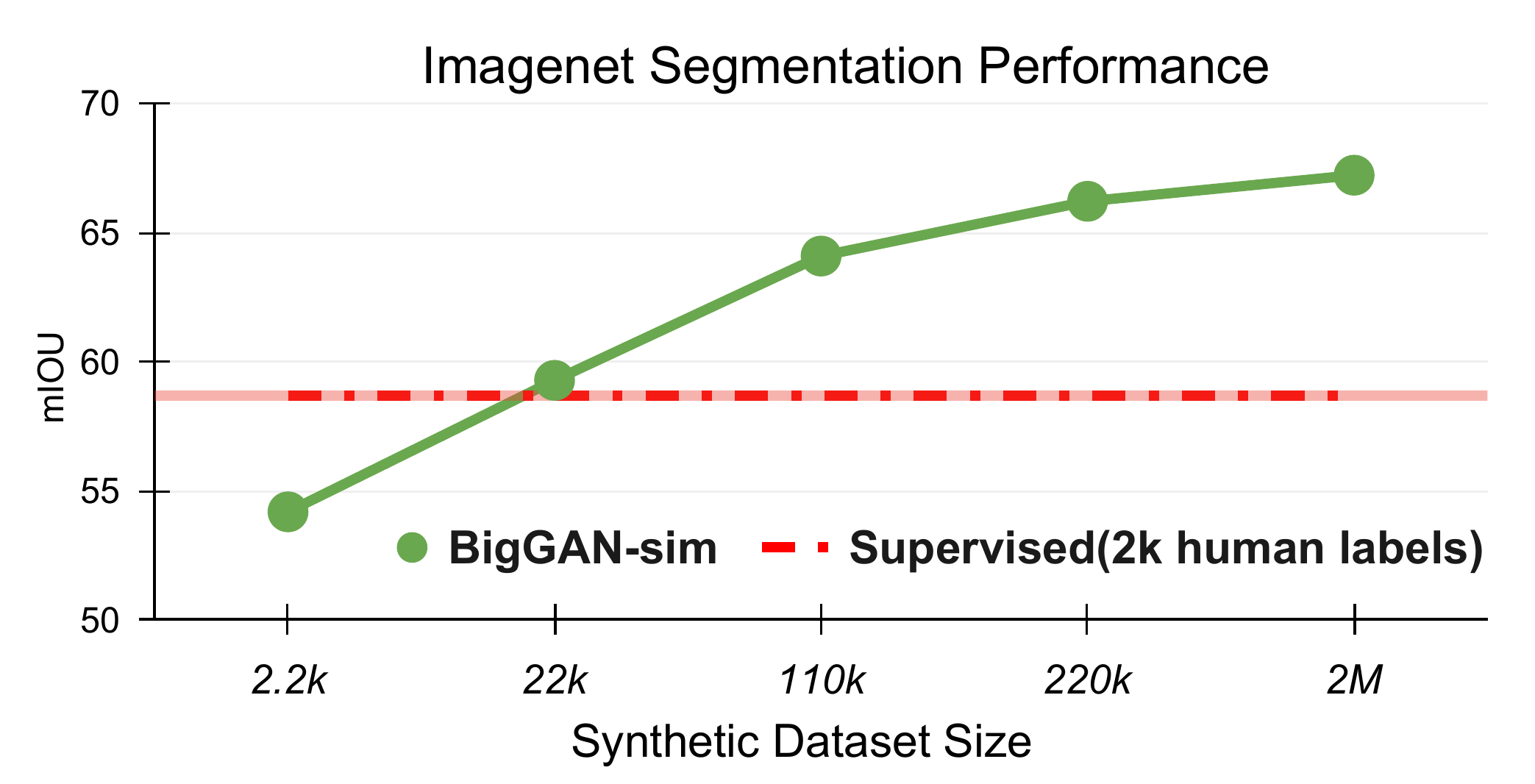}
\end{center}
\vspace{-6mm}
\caption{
\footnotesize\textbf{Ablating synthetic dataset size.}
Here we fix the model to the Resnet50 backbone and compare the performance when we increase the synthetic dataset size. The model trained using a 22k synthetic dataset outperforms the same model trained with 2k human-annotated dataset. Another 7 points is gained when further increasing the synthetic data size from 22k to 220k. Here, 2M is the total number of samples synthesized through our online sampling strategy (See Sec.~\ref{sec:synth_dataset}).
}
\label{fig:datasize}
\end{minipage}
\hfill
\begin{minipage}{0.32\textwidth}
\begin{center}
\includegraphics[width=.99\columnwidth,trim=0 0 0 30,clip,height=3.3cm]{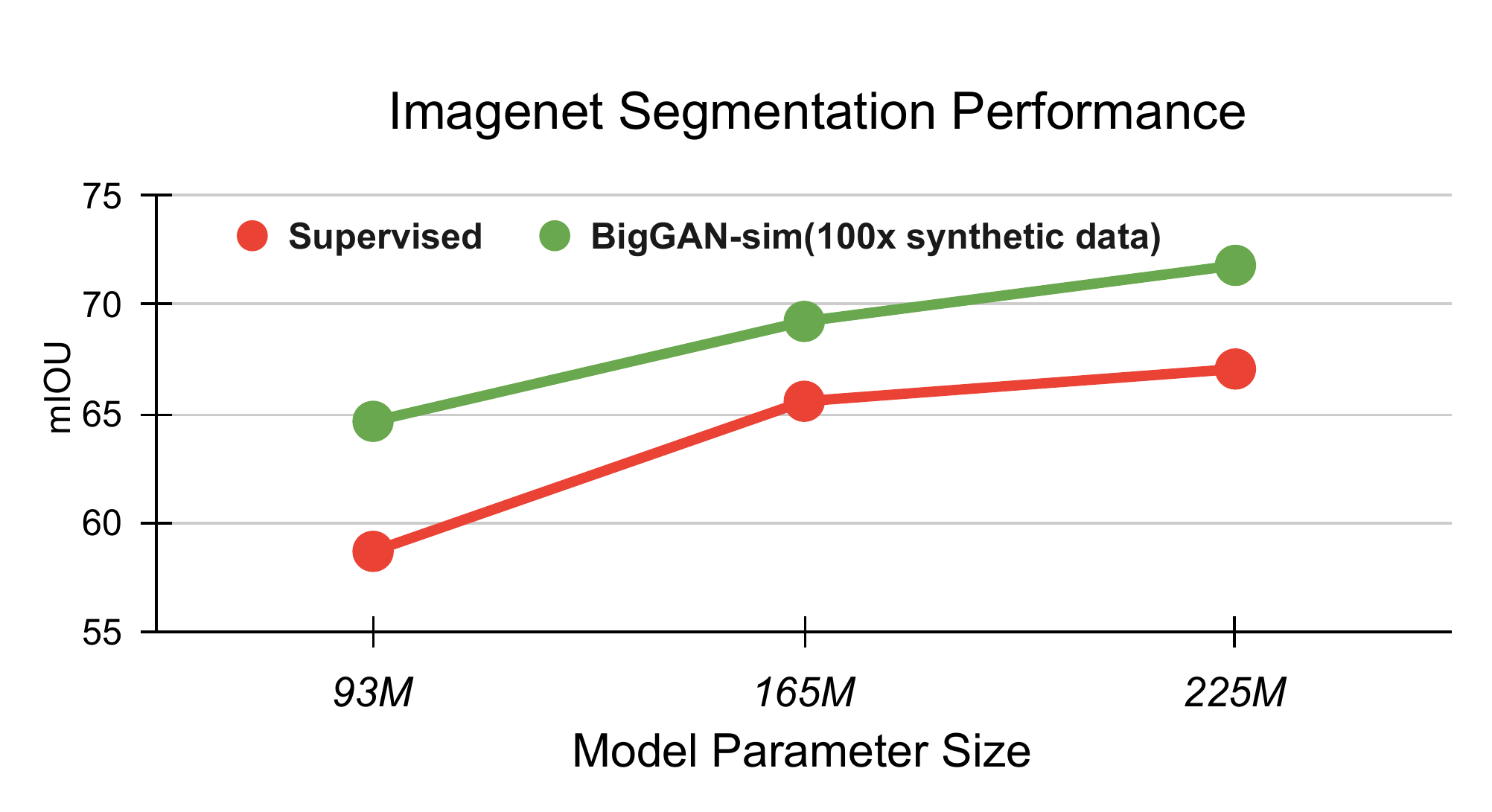}
\end{center}
\vspace{-7mm}
\caption{
\footnotesize \textbf{Ablating backbone size.}
We scale up the backbone from Resnet50 to Resnet101 and Resnet152.
We supervise with 2k human-annotated labels \textit{(red)}, and with our BigGAN-sim dataset \textit{(green)}, which is 100x larger. BigGAN-sim dataset supervision leads to consistent improvements, especially for larger models.
}
\label{fig:modelsize}
\end{minipage}
\hfill
\begin{minipage}{0.32\textwidth}
\begin{center}
\includegraphics[width=.99\linewidth,trim=20 5 20 30,clip]{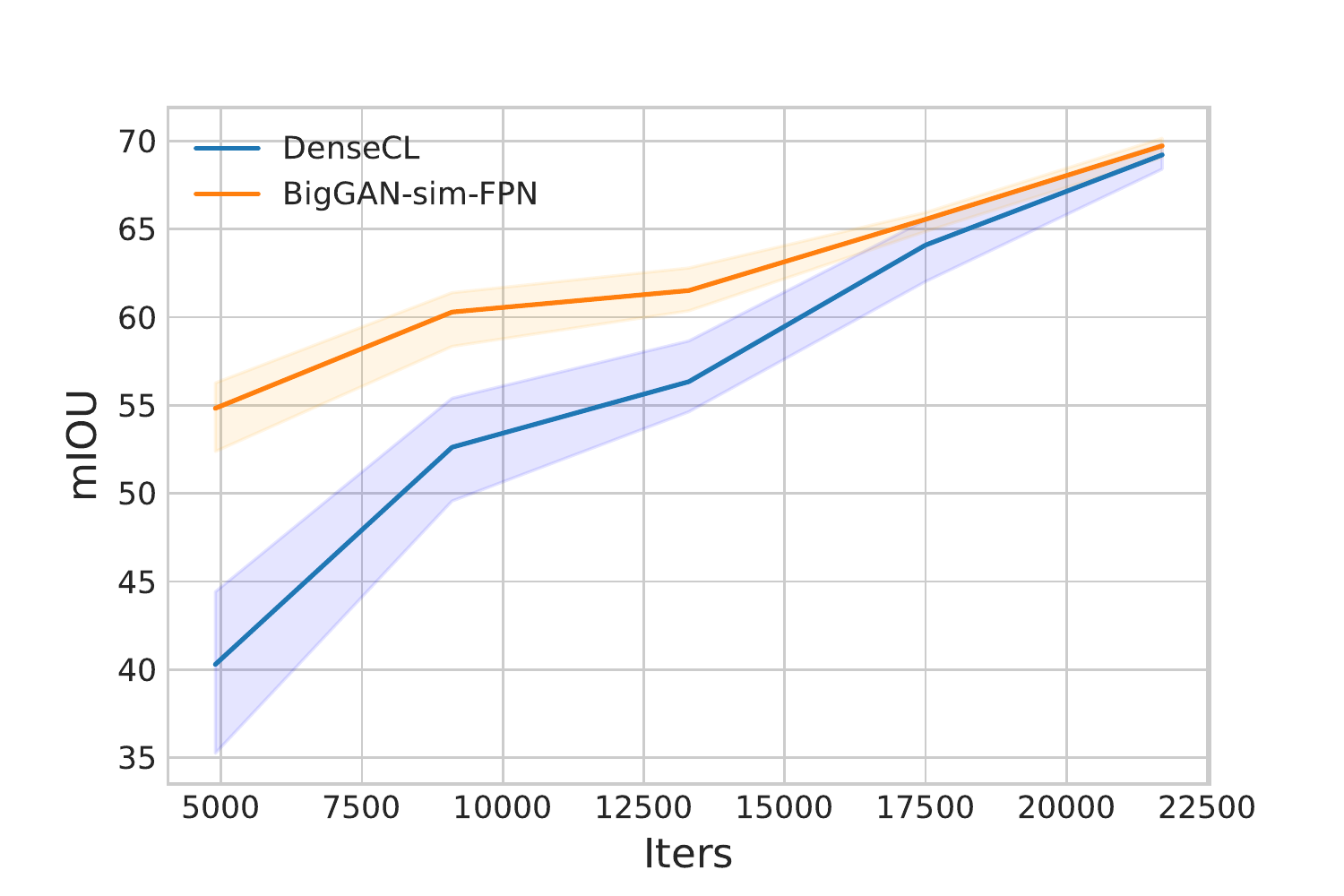}
\end{center}
\vspace{-6mm}
\caption{
\footnotesize\textbf{Convergence of segm. performance on Pascal-VOC.} We evaluate val. set semantic segmentation at different training iterations. Backbone pre-trained using our synthetic dataset shows significantly faster convergence rates compared to training with contrastive learning alone. Colored area indicates variance over 5 trials.
}
\label{fig:voc-seg}
\end{minipage}
\vspace{-2mm}
\end{figure*}

%% file: fig/benchmark_testing/benchmark_testing.tex
\newcommand\hh{1.45cm}
\newcommand\ww{4.34cm}


\begin{figure*}[t!]
\vspace{-0mm}
\addtolength{\tabcolsep}{-5.5pt}
\begin{tabular}{cccc}

\includegraphics[height=\hh,width=\ww, trim=0 0 0 0,clip]{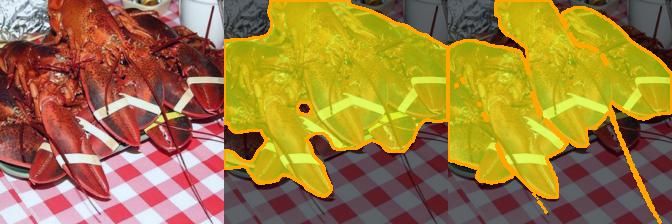} & \includegraphics[height=\hh,width=\ww, trim=0 0 0 0,clip]{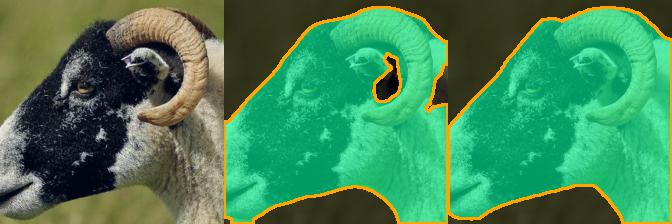} & \includegraphics[height=\hh,width=\ww,  trim=0 0 0 0,clip]{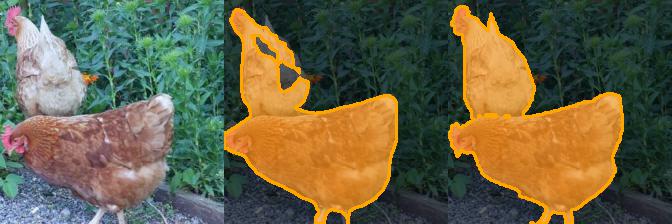} &
\includegraphics[height=\hh,width=\ww,  trim=0 0 0 0,clip]{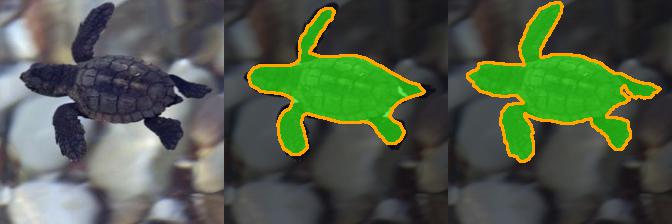}
 \\
\includegraphics[height=\hh,width=\ww, trim=0 0 0 0,clip]{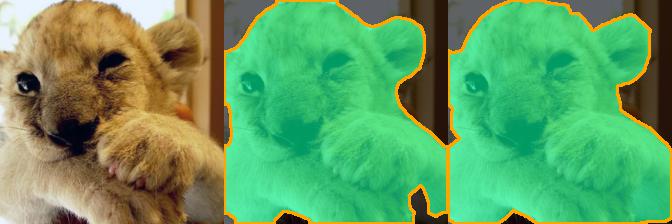} & \includegraphics[height=\hh,width=\ww, trim=0 0 0 0,clip]{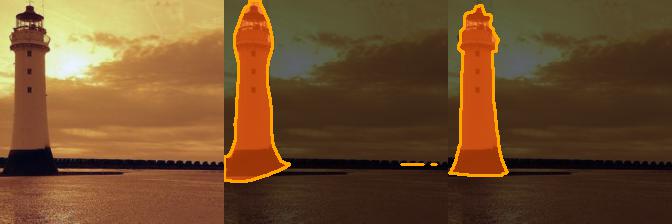} & \includegraphics[height=\hh,width=\ww,  trim=0 0 0 0,clip]{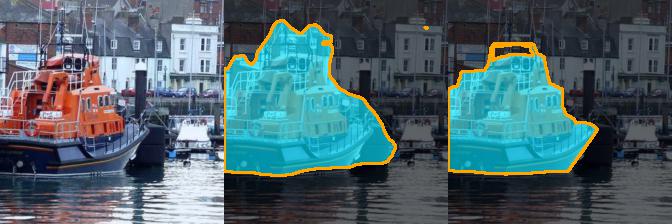} &
\includegraphics[height=\hh,width=\ww,  trim=0 0 0 0,clip]{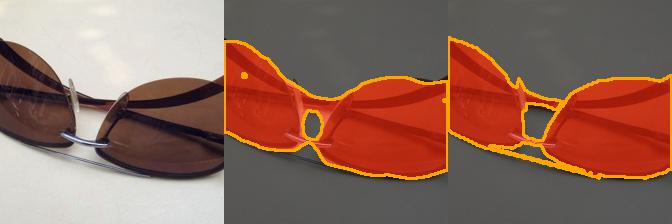} 
 \\
 
 \includegraphics[height=\hh,width=\ww, trim=0 0 0 0,clip]{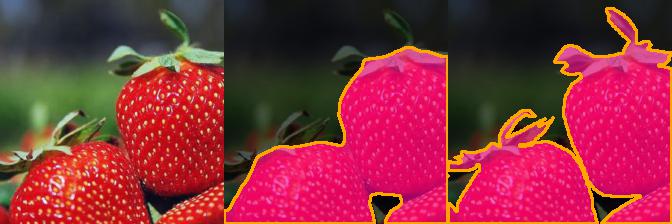} & \includegraphics[height=\hh,width=\ww, trim=0 0 0 0,clip]{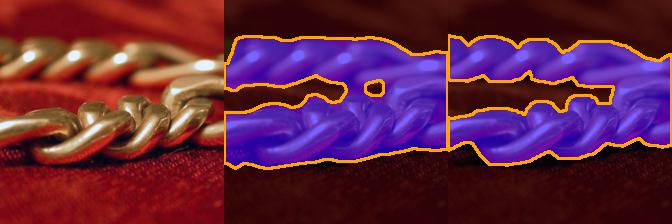} & \includegraphics[height=\hh,width=\ww,  trim=0 0 0 0,clip]{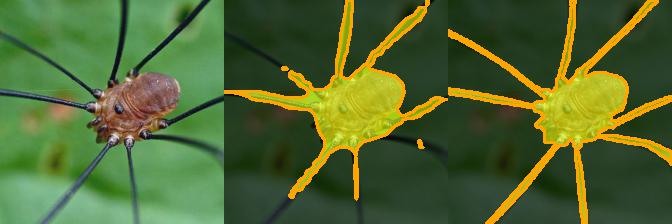} &
\includegraphics[height=\hh,width=\ww,  trim=0 0 0 0,clip]{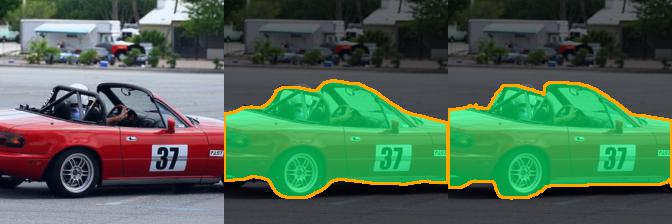}
 \\
 
  \includegraphics[height=\hh,width=\ww, trim=0 0 0 0,clip]{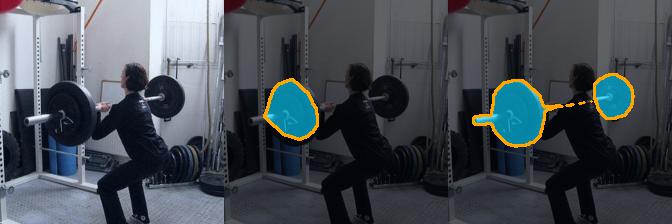} & 
  \includegraphics[height=\hh,width=\ww, trim=0 0 0 0,clip]{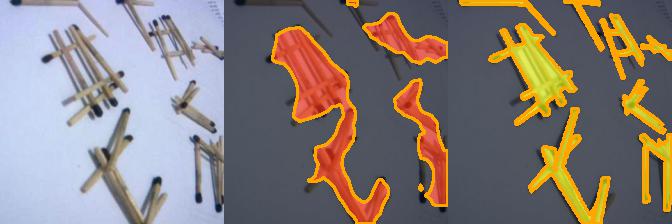} & \includegraphics[height=\hh,width=\ww,  trim=0 0 0 0,clip]{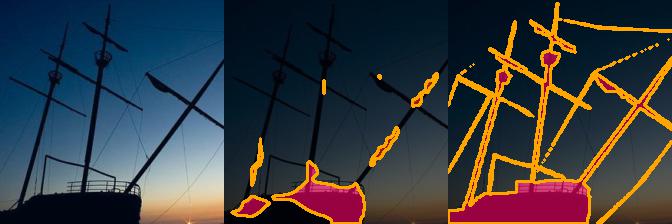} 
  &
\includegraphics[height=\hh,width=\ww,  trim=0 0 0 0,clip]{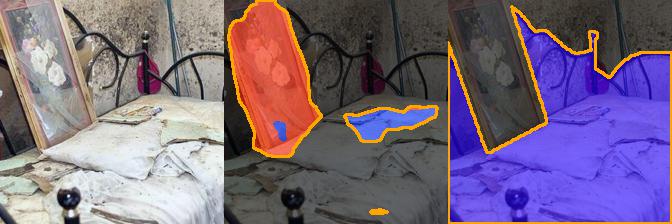}
 \\

\end{tabular}
\vspace{-4mm}
\caption{\footnotesize  {\bf Qualitative results on MC-128.} We visualize predictions (\textit{second} column) of DeepLab trained on our BigGAN-sim dataset, compared to ground-truth annotations (\textit{third} column). The final row shows typical failure cases, which include multiple parts, thin structures or complicated scenes.}
\label{fig:vis_testing}
\vspace{-2mm}
\end{figure*}

%% file: tab/mscoco.tex
\begin{table*}[t]
    \begin{subtable}[h]{0.49\textwidth}
        \centering
        \resizebox{\linewidth }{!}{ 
        \begin{tabular}{c|ccc|ccc}
        
        pre-train & $AP^{bb}$ & $AP^{bb}_{50}$ & $AP^{bb}_{75}$ & $AP^{mk}$ & $AP^{mk}_{50}$ & $AP^{mk}_{75}$ \\
        \hline
        random init. & 31.0 & 46.8 & 30.4 & 28.5 & 46.8 & 30.4 \\
        supervised   & 38.9 & 59.6 & 42.7 & 35.4 & 56.5 & 38.1 \\
        \hline
        MoCo-v2\cite{chen2020improved}    & 38.9 & 59.2 & 42.4 & 35.4 & 56.2 & 37.8 \\
        DenseCL\cite{wang2021dense}      & 39.4 & 59.9 & 42.7 & 35.6 & 56.7 & 38.2 \\
        \hline
        BigGAN-sim-FPN & \textbf{39.8} & \textbf{60.4} & \textbf{43.7} & \textbf{35.9} & \textbf{57.3} & \textbf{38.2} \\
        \end{tabular}}
        \caption{Mask R-CNN R50-FPN, 1× schedule}
    \end{subtable}
    \hfill
    \begin{subtable}[h]{0.49\textwidth}
        \centering
        \resizebox{\linewidth }{!}{ 
        \begin{tabular}{c|ccc|ccc}
        
        pre-train & $AP^{bb}$ & $AP^{bb}_{50}$ & $AP^{bb}_{75}$ & $AP^{mk}$ & $AP^{mk}_{50}$ & $AP^{mk}_{75}$ \\
        \hline
        random init. & 36.7 & 56.7 & 40.0 & 33.7 & 53.8 & 35.9 \\
        supervised   & 40.6 & 61.3 & 44.4 & 36.8 & 58.1 & 39.5 \\
        \hline
        MoCo-v2\cite{chen2020improved}  & 40.9 & 61.5 & 44.6 & 37.0 & 58.4 & 39.6 \\
        DenseCL\cite{wang2021dense}  & 41.2 & 61.9 & 45.1 & 37.3 & 58.9 & 40.1 \\
        \hline
        BigGAN-sim-FPN & \textbf{41.5} & \textbf{62.2} & \textbf{45.4} & \textbf{37.5} & \textbf{59.0} & \textbf{40.5} \\
        \end{tabular}}
        \caption{Mask R-CNN R50-FPN, 2× schedule}
    \end{subtable}
    
\vspace{-3mm}
\caption{
\textbf{MS-COCO object detection \& instance segmentation.} Using our synthetic data during pre-training improves object detection performance by 0.4 $AP^{bb}$, and instance segmentation by 0.3 $AP^{mk}$ in $1\times$ training schedule. When training longer in the $2\times$ schedule, our synthetic data consistently helps improving the task performance by 0.3 $AP^{bb}$ and 0.2 $AP^{mk}$.
} 
\label{tab:mscoco-det}
\end{table*}

%% file: tab/voc-det.tex
\begin{table}[t]

\centering
\resizebox{\linewidth }{!}{ 
\setlength{\tabcolsep}{12pt}
\begin{tabular}{c|ccc|c}
& \multicolumn{3}{c}{\textbf{Detection}} & \multicolumn{1}{c}{\textbf{Semantic seg.}} \\
pre-train & $AP$ & $AP_{50}$ & $AP_{75}$ & \textit{mIoU}\\
\hline
random init. & 33.8 & 60.2 & 33.1 & 40.7\\
supervised   & 53.5 & 81.3 & 58.8 & 67.7\\
\hline
MoCo-v2\cite{chen2020improved}  & 57.0 & 82.4 & 63.6 & 67.5\\
DenseCL\cite{wang2021dense} & \textbf{58.7} & 82.8 & 65.2 & 69.2\\
\hline
BigGAN-sim-FPN & 58.2 & \textbf{83.0} & \textbf{65.5} & \textbf{69.7}\\

\end{tabular}}
\vspace{-3mm}
\caption{
\textbf{PASCAL VOC detection \& semantic segmentation.} For detection, we follow~\cite{he2020momentum} and train on the \texttt{trainval'07+12} set and evaluate on \texttt{test07}. For semantic segmentation, we train on \texttt{train aug2012} \cite{li2021semantic} and evaluate on \texttt{val2012}. Results are average over 5 individual trials.
} 
\label{tab:voc-det}
\end{table}

%% file: tab/cxr-seg.tex
\begin{table}[t]
\small
\centering
\resizebox{0.95\linewidth }{!}{ 
\setlength{\tabcolsep}{6pt}
\begin{tabular}{c|ccccc}

pre-train & $1\%$ & $5\%$ & $10\%$ & $50\%$ & $100\%$ \\
\hline
supervised                       & 49.6 & 47.8 & 54.1 & 57.6 & 63.6 \\
DenseCL\cite{wang2021dense}      & 48.8 & 59.4 & 66.2 & 69.2 & 70.4 \\
\hline
BigGAN-sim-FPN & \textbf{67.6} & \textbf{70.8} & \textbf{71.2} & \textbf{74.7} & \textbf{75.3} \\

\end{tabular}}
\vspace{-3mm}
\caption{
\textbf{Semi-supervised chest X-ray segmentation with a frozen backbone}. Performance numbers are mIoU. When using our synthetic dataset, we match the performance of the supervised and self-supervised pre-trained networks with only 1\% and 5\% of labels, respectively. We achieve a big gain using 100\% of the data. Numbers are averaged over 3 independent trials. 
} 
\label{tab:cxr-seg}
\end{table}

%% file: tab/cityscape-voc.tex
\begin{table}[t]

\centering
\resizebox{\linewidth }{!}{ 
\setlength{\tabcolsep}{12pt}
\begin{tabular}{l|cc|c}
& \multicolumn{2}{c}{\textbf{Instance seg.}} & \multicolumn{1}{c}{\textbf{Semantic seg.}} \\
pre-train & $AP^{mk}$ & $AP^{mk}_{50}$ & \textit{mIoU} \\
\hline
random init. & 25.4 & 51.1 & 63.5  \\
supervised   & 32.9 & 59.6 & 73.7 \\
\hline
MoCo-v2\cite{chen2020improved}  & 33.7 & 61.8 & 74.5 \\
DenseCL\cite{wang2021dense} & 34.5 & 62.1 & {\bf 75.7} \\
\hline
BigGAN-sim-FPN & {\bf 34.8} & {\bf 62.9} & 75.6\\
\end{tabular}}
\vspace{-3mm}
\caption{
\textbf{Cityscapes instance and semantic segmentation.} We train on \texttt{train\_fine} set and evaluate on \texttt{val} set.
} 
\label{tab:cityscape-voc}
\vspace{-4mm}
\end{table}

%% file: sec/5_conclusions.tex
\section{Conclusions}

In this paper, we collected a new ImageNet benchmark targeting segmentation of all 1k classes. For scalability, we labeled a modestly sized dataset of images sampled from an ImageNet-trained BigGAN model.
By enhancing this BigGAN and another ImageNet-trained VQGAN with segmentation branches 
we obtain high quality dataset generators, named \textit{BigDatasetGAN}. Our BigDatasetGAN synthesize large-scale, densely labeled datasets, which 
we show are highly beneficial for a variety of downstream tasks when combined with different supervised or self-supervised pretrained
backbone models.
Due to the scalability of our approach, which affords a minimal labeling effort, we aim to extend our ImageNet annotation to include object parts and other dense labels, such as object keypoints, in the future. 

%% file: sec/X_supplementary.tex
\appendix


\twocolumn[
\centering
\Large
\textbf{BigDatasetGAN: Synthesizing ImageNet with Pixel-wise Annotations} \\
\vspace{0.5em}Supplementary Material \\
\vspace{1.0em}
] 
\appendix

\section{BigDatasetGAN Implementation Details}
\subsection{BigGAN}
\vspace{-3mm}
\paragraph{Training}
When training the segmentation branch of BigGAN, we first load the pre-recorded latent codes of the annotated samples. We then pass the latent codes as well as the class information into the pre-trained BigGAN model (with resolution $512\times512$) to extract the features. With BigGAN features and class information (input to BigGAN), we can train the segmentation branch with the binary cross-entropy loss on the human-annotated segmentation masks. Note that in this case we exploit the fact that only foreground object and background scene are annotated in each image and the class information is known. 
We use the Adam~\cite{kingma2017adam} optimizer with a learning rate $0.001$ and batch size of 8. We train the segmentation branch with 5k iterations.
\vspace{-2mm}

\paragraph{Uncertainty Filtering}
As mentioned in Sec 3.2 in the main paper, we apply uncertainty filtering on top of the truncation trick~\cite{brock2019large} and rejection sampling~\cite{razavi2019generating}. We train 16 segmentation heads, and follow~\cite{kuo2018costsensitive} and~\cite{zhang2021datasetgan} to use the Jensen-Shannon (JS) divergence as the uncertainty measure. 
Concretely, the JS divergence is defined as:
\begin{equation*}
    JS(P_{1},P_{2},...,P_{N}) = H\left(\frac{1}{N}\sum_{i}^{N} P_{i}\right) - \frac{1}{N}\sum_{i}^{N}H(P_i),
\end{equation*}
where $N$ is the number of models, in our case is 16. $H$ is the entropy function and $P_i$ is the output distribution of model $i$. We record the uncertainty score for each sample and then follow~\cite{zhang2021datasetgan} to filter out the $10\%$ most uncertain image samples.

\subsection{VQGAN}
\vspace{-3mm}
\paragraph{Architecture Details}
We use VQGAN~\cite{esser2020taming} trained on ImageNet at $256\times256$ resolution.
VQGAN's class-conditional transformer consists of 48 self-attention~\cite{vaswani2017attention} layers, each with 1536 dimensions, operating on the $16\times16$ codebook with the vocabulary size of 16384.
We gather features from every fourth transformer layer for each spatial location ($16\times16$) of the encoder output.
Additionally, we gather features from the decoder layers at $16\times16$ to $256\times256$ resolutions.
In total, $F_\textrm{VQGAN}$ consists of 12 transformer features and 7 decoder features.
Similarly to BigGAN, we group features according to their spatial resolutions into a high-level group ($16\times16$ to $32\times32$), a mid-level group ($64\times64$ to $128\times128$), and a low-level group ($256\times256$).
We reduce the dimensionality of each feature to 128 using a \texttt{1x1conv} operation and then \texttt{upsample} features within the same group into the highest resolution within the group.
Features from two separate levels are fused by upsampling the features of the previous, higher level group to match the current layer's spatial resolution and then concatenating the two levels, followed by a \texttt{mix-conv} operation which contains a \texttt{3x3conv} layer.
Finally, three \texttt{1x1conv} layers are used to output the segmentation logits. We use layer normalization~\cite{ba2016layer} between the \texttt{1x1conv} layers.
\vspace{-2mm}

\paragraph{Training}
To train the segmentation branch for VQGAN, we leverage its encoder to encode the images that were previously obtained via BigGAN sampling (with resolution $256\times256$). We can then extract the features as described above and obtain
the segmentation logits.
Additionally, the class-conditional transformer takes in the class label as input.
We use the same binary cross-entropy loss for the foreground and background segmentation as in BigGAN training. 
We use the Adam~\cite{kingma2017adam} optimizer with a learning rate of 0.001 and batch size of 8. 


\section{ImageNet Segmentation Benchmark Details}
\input{tab/benchmark-split}
\subsection{Dataset Statistics}
Table~\ref{tab:benchmark-split} provides the dataset split details for the 7 tasks in the benchmark. Please refer to Sec 5.1 in the main paper for the dataset details and Sec 5.2 for details about benchmark tasks. We plan to provide a detailed class mapping when launching the public benchmark.

\subsection{Training Setup}
For all the baselines and our methods, we use DeepLab-v3~\cite{deeplabv32018} with Resnet-50 \cite{He2015} as the image backbone model. We use the SGD optimizer with learning rate $0.01$, momentum $0.9$ and weight-decay $0.0001$. We use polynomial learning rate decay with power $0.9$. We use batch size 64 for all tasks and train for 200 epochs. For augmentation, we use \textit{random resize} with scales from $0.5-2.0$. \textit{random crop} and \textit{random horizontal filp}. We use resolution 224 for both training and testing. When training with our synthetic dataset, we use the same training schedule and augmentation policy as described here. 

As objective, we use the cross-entropy loss function for all tasks except for \textit{MC-992}. The task \textit{MC-992} needs to segment over 992 classes, where the object pixel distribution is varying greatly between different classes. This corresponds to an imbalanced training setup, which makes learning the model difficult. We found that training did not converge well using the standard cross-entropy loss. In order to mitigate this issue, we use the focal loss~\cite{lin2018focal} for all the methods in task \textit{MC-992}.

\input{tab/imagenet-seg-online}
\subsection{Long-time Training with Online Sampling}
Since the online sampling strategy needs to do a forward pass through the generative model to sample labeled examples at each iteration, the training is slower compared to offline sampling. Because of that, the result reported in Table 2 in the main paper for \textit{BigGAN-on} on the \textit{MC-992} task corresponds to a training run that is not fully converged. Here, we report updated results for \textit{BigGAN-on} in Table~\ref{tab:imagenet-seg-online}.

\section{Dataset Analysis Details}
\input{fig/supp/center_scatter_plot}
\subsection{Center Distributions}
We visualize scatter plots of object center locations for the \textit{Real-annotated}, \textit{Synthetic-annotated}, \textit{BigGAN-sim} and \textit{VQGAN-sim} datasets in Figure~\ref{fig:center-scatter}. We fit a tight bounding box over the segmentation mask and use the center of the bounding box as the center location. Note that the location is normalized with respect to the original image size. We see that most of the object centers are biased towards the image center for all datasets. Compared to the human-annotated datasets, \textit{BigGAN-sim} and \textit{VQGAN-sim} show noisier center distributions, in particular towards the image corners.
\subsection{Geometry Metrics}
In Table 1 in the main paper, we compute geometry metrics to measure shape complexity (SC) and shape diversity (SD) of the human-annotated dataset as well as the synthetic dataset. Here, we provide the corresponding implementation details.

\paragraph{Implementation.} We first filter out masks with a total number of pixels below 100 to avoid noisy labels. We then calculate the connected components in the mask and only use the largest connected component in our measurement. We further use OpenCV's \texttt{findContours} function with the \texttt{CHAIN\_APPROX\_SIMPLE} flag which compresses horizontal, vertical, and diagonal segments and leaves only their end points to extract a simplified polygon. Note that the simplification happens in pixel space and the extracted polygons do not have the same scales. In order to overcome such scaling issues, we normalize the extracted polygon to a unit square by $p_{i} = (p_{i} - p_{min}) / (p_{max} - p_{min})$ (this operation is applied separately for both horizontal and vertical directions in an image),
where $p_{i}$ is the point in the extracted polygon, and $p_{min}$ and $p_{max}$ are the minimum and maximum point coordinates (for either horizontal or vertical direction) over the set of points for a given polygon. We further apply the Douglas-Peucker algorithm~\cite{douglas1973algorithms} with a threshold of $0.01$ to remove redundant points.

\paragraph{Shape Complexity} After applying the pre-processing steps mentioned above, the shape complexity (SC) is calculated as the number of points of the normalized simplified polygon. We also measure polygon length (PL) in Table 1 from the main paper.

\paragraph{Shape Diversity} We calculate the shape diversity by mean pair-wise Chamfer distance~\cite{fan2016point} per class and average across classes. Specifically, Chamfer distance is defined as 
\begin{equation*}
    d_\textrm{CD}(S_1,S_2) = \sum_{\mathbf{x} \in S_1} \min_{\mathbf{y} \in S_2} ||\mathbf{x}-\mathbf{y}||^2_2 + \sum_{\mathbf{y} \in S_s} \min_{\mathbf{x} \in S_1} ||\mathbf{x}-\mathbf{y}||^2_2
\end{equation*}
where $S_1$ and $S_2$ are sets of points corresponding to different polygons. In our case, we calculate average pair-wise $d_\textrm{CD}$ between all polygons within each class. Then, we compute the shape diversity (SD) metric as the average distance over all classes.

\section{Transfer Learning Experiments}
\subsection{Pre-training Implementation}
We closely follow the pre-training setup of DenseCL~\cite{wang2021dense}, and use SGD as the optimizer with weight decay and momentum set to $0.0001$ and $0.9$, respectively. The dictionary size is set to 65536 and the momentum update rate of the encoder is set to 0.999. We also adopt its augmentation policy with $224 \times 224$ random resized cropping, random color jittering, random gray-scale conversion, Gaussian blurring and random horizontal flips for contrastive learning. Note that since texture and color are important hints for semantic segmentation, we only use random resized cropping with resolution 224 $\times$ 224 and random horizontal flip when training the segmentation branch with our synthetic dataset. The batch size is 256, and we train on 8 GPUs with a cosine learning rate decay schedule. We pre-train the backbone on ImageNet using the original contrastive losses for 150 epochs and then jointly train the segmentation branch using the cross-entropy loss for another 50 epochs.

\subsection{Downstream Task Details}
\vspace{-3mm}
\paragraph{Benchmark Implementation} For object detection and instance segmentation tasks on Pascal VOC, MS-COCO and Cityscapes, we use standard benchmark configurations from OpenSelfSup\footnote{https://github.com/open-mmlab/OpenSelfSup}. The model is trained using Detectron2~\cite{wu2019detectron2}. For semantic segmentation tasks on Pascal VOC and Cityscapes, we use configurations from DenseCL\footnote{https://github.com/WXinlong/DenseCL} implemented in MMSegmentation~\cite{mmseg2020}.

\section{Qualitative Results}
We show random samples from the human-annotated and synthetic datasets. See Figure~\ref{fig:real-samples} for dataset samples from the \textit{Real-annotated} dataset where annotation is done on real images. Figure \ref{fig:sim-samples} shows dataset samples from the \textit{Synthetic-annotated} dataset where human annotation is on BigGAN generated images. For our synthetic dataset \textit{BigGAN-sim} generated by BigGAN, see Figure \ref{fig:biggan-random-samples}, and for the \textit{VQGAN-sim} dataset generated by VQGAN, see Figure \ref{fig:vqgan-random-samples}. Compared to the synthetic dataset, \textit{Real-annotated} images have more complex backgrounds and structure. However, we also see that the synthetic datasets generated by BigGAN and VQGAN include photo-realistic and diverse images as well as high quality labels.

We also show per-class samples where images in the same row are from the same class. For \textit{BigGAN-sim} per-class samples, please see Figure \ref{fig:biggan-perclass-samples}. For \textit{VQGAN-sim} per-class samples, see Figure \ref{fig:vqgan-perclass-samples}. Note that we select the same classes for both \textit{BigGAN-sim} and \textit{VQGAN-sim} for easy comparison. Comparing to \textit{BigGAN-sim}, the \textit{VQGAN-sim} dataset samples are more diverse in terms of object scale, pose as well as background. However, we see \textit{BigGAN-sim} has better label quality than \textit{VQGAN-sim} where in some cases the labels have holes and are noisy. 

We also include mean shape visualizations for different classes from the \textit{BigGAN-sim} dataset (see Figure \ref{fig:biggan-meanshapes}). We first use a tight bounding box to fit the segmentation, and then crop and resize the segmentation into resolution $32 \times 32$. We use k-means clustering with $k=5$ for each class to calculate the major modes of the resized segmentation mask. In Figure \ref{fig:biggan-meanshapes}, we randomly select 500 classes and for each class we randomly select one cluster out of the 5 clusters and visualize the mean shape. We see diverse shapes with different poses especially for classes related to animals. Some classes do not have clear shapes. This might be because the randomly selected mode is potentially not the most meaningful mode.

\input{fig/supp/real-annotated-samples}
\input{fig/supp/sim-annotated-samples}
\input{fig/supp/biggan-sim-random-samples}
\input{fig/supp/vqgan-sim-random-samples}
\input{fig/supp/biggan-sim-perclass-samples}
\input{fig/supp/vqgan-sim-perclass-samples}
\input{fig/supp/biggan-sim-meanshapes}

%% file: tab/benchmark-split.tex
\begin{table}[t]

\centering
\resizebox{\linewidth }{!}{ 
\renewcommand{\arraystretch}{1.2}
\setlength{\tabcolsep}{6pt}
\begin{tabular}{l|c|c|c|c|c|c|c}
 & \textit{Dog} & \textit{Bird} & \textit{FG/BG} & \textit{MC-16} & \textit{MC-100} & \textit{MC-128} & \textit{MC-992} \\
\hline
train & 657 & 366 & 5294 & 1268 & 540 & 5294 & 5294 \\
test  & 1040 & 512 & 8316 & 1967 & 798 & 8316 & 8316 \\

\end{tabular}}
\vspace{-3mm}
\caption{
\textbf{ImageNet Segmentation Benchmark Splits.} The training set is based on \textit{Synthetic-annotated} (Images sampled from BigGAN), while the testing set consists of images from \textit{Real-annotated}. 
} 
\label{tab:benchmark-split}
\vspace{-4mm}
\end{table}

%% file: tab/imagenet-seg-online.tex
\begin{table}[t]
\centering
\resizebox{1.00\linewidth }{!}{ 
\setlength{\tabcolsep}{2pt}
\renewcommand{\arraystretch}{1.1}

\hspace{-2.5mm}\begin{tabular}{c|l|c|c|c|c|c|c|c||c}
& Method  & \textit{Dog} & \textit{Bird} & \textit{FG/BG} & \textit{MC-16} & \textit{MC-100} & \textit{MC-128} & \textit{MC-992} & Mean\\
\hline
\parbox[t]{3.1mm}{\multirow{6}{*}{\rotatebox[origin=c]{90}{superv. pre-train.}}} & Rand                                        & 56.4 & 35.7 & 44.7 & 13.9 & 4.0 & 3.6 & 2.3 & 22.9\\
\cdashline{2-10} 
& Sup.IN                                      & 82.6 & 79.6 &  66.3 & 58.7 & 56.1 & 28.5 & 17.8 & 55.7\\
& $\ $ +  BigGAN-off                             & 85.8 & 81.2 & 67.5 & 64.6 & 62.3 & 29.3 & 22.8 & 59.0\\
& $\ $ +  BigGAN-on                               & \textbf{87.0} & \textbf{83.2} & \textbf{69.5} & \textbf{66.1} & \textbf{62.8} & \textbf{29.5} & \textbf{24.6} & \textbf{60.4}\\
\cdashline{2-10} 
 & SupCon~\cite{khosla2021supervised}           & 83.8 & 79.0 & 66.6 & 59.2 & 55.4 & 28.6 & 18.7 & 55.9\\
& SupSelfCon~\cite{islam2021broad} & 84.4 & 81.8 & 67.6 & 63.1 & 60.0 & 28.3 & 18.9 & 57.8\\
& $\ $ +  BigGAN-sim                              & \textbf{87.0} & 83.2 & 69.5 & 66.1 & 62.8 & \textbf{32.8} & \textbf{29.7} & \textbf{61.6}\\
& $\ $ +  VQGAN-sim                               & 86.7 & \textbf{84.4} & \textbf{71.1} & \textbf{68.1} & \textbf{64.7} & 30.4 & 25.8 & \textbf{61.6}\\

\end{tabular}
} 
\vspace{-3mm}
\caption{
\textbf{ImageNet pixel-wise benchmark.} Here, we include supervised pre-training results for our benchmark, similar to Table 2 in the main paper. We only updated the results for the \textit{BigGAN-on} method on task \textit{MC-992}, since the number reported in the main paper corresponds to a not fully converged training run.
} 
\label{tab:imagenet-seg-online}
\vspace{-4mm}
\end{table}

%% file: fig/supp/center_scatter_plot.tex
\begin{figure}[t]
\begin{center}
\includegraphics[width=1.0\linewidth]{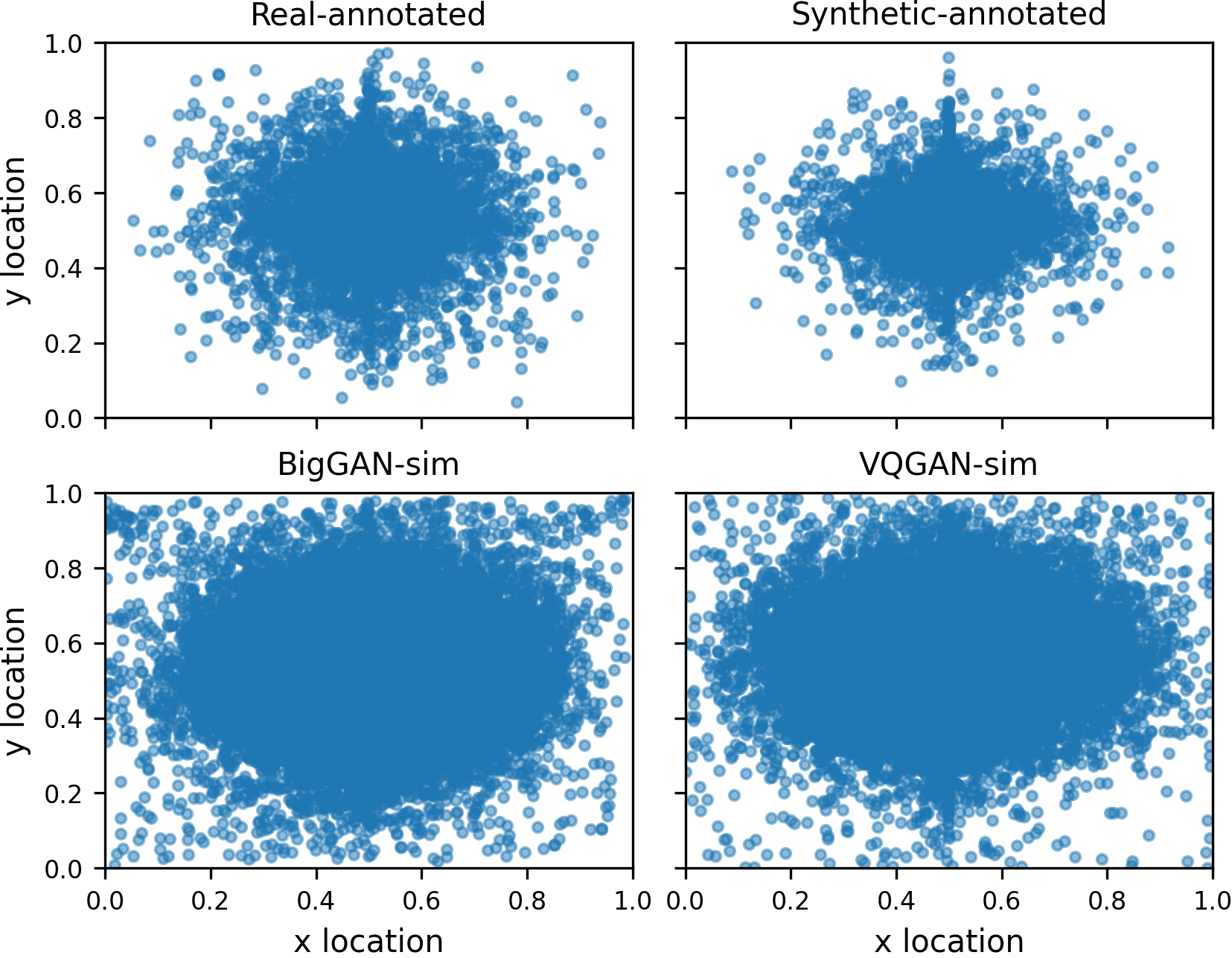}
\end{center}
\vspace{-3mm}
\caption{
\textbf{Object center scatter plots.} We visualize center positions of the tight bounding box of pixel-wise labels for the \textit{Real-annotated}, \textit{Synthetic-annotated}, \textit{BigGAN-sim} and \textit{VQGAN-sim} datasets. The box center is normalized with respect to the original image size.
}
\label{fig:center-scatter}
\vspace{-3mm}
\end{figure}

%% file: fig/supp/real-annotated-samples.tex
\begin{figure*}[t!]
\begin{center}
\includegraphics[width=1.0\linewidth]{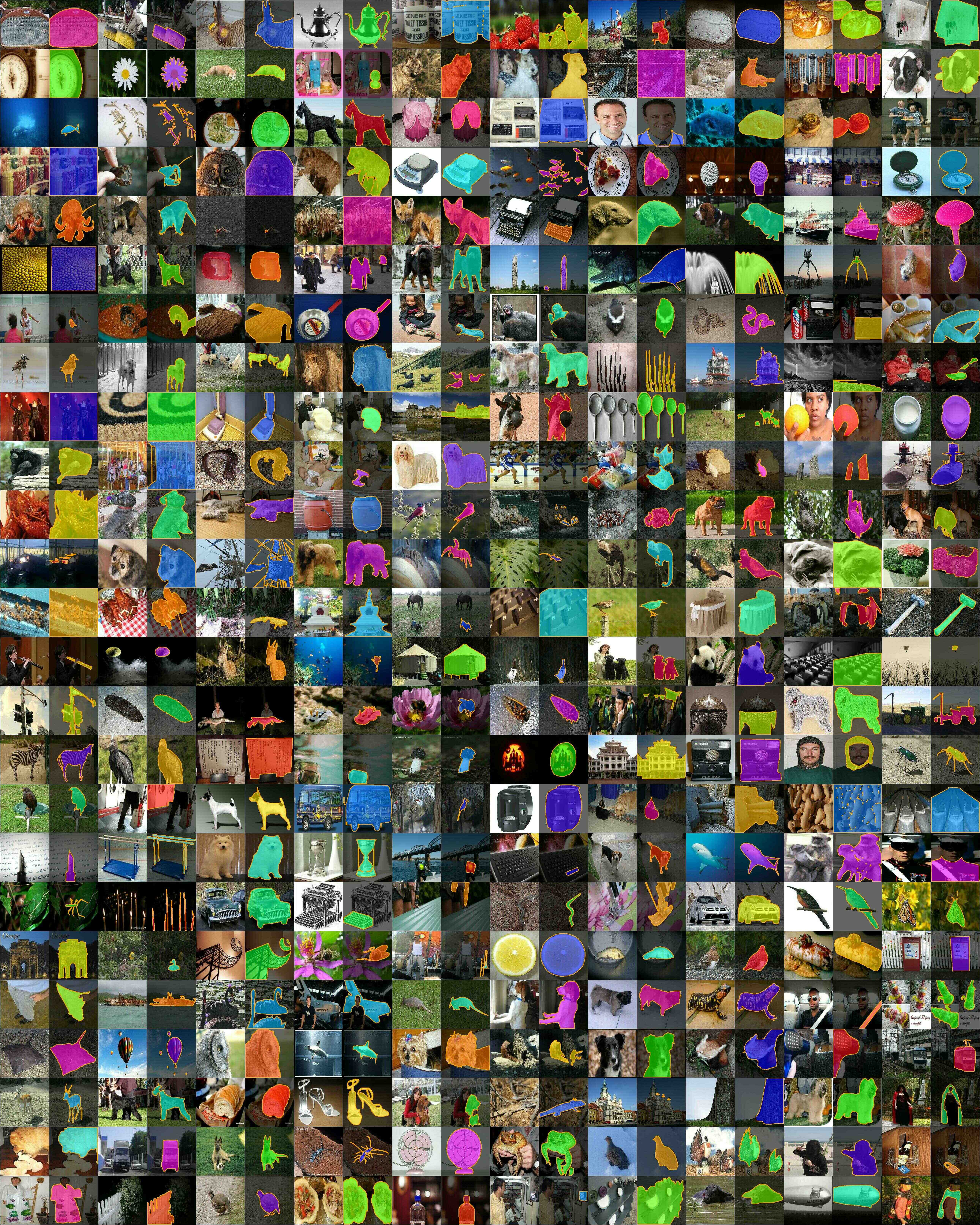}
\end{center}
\vspace{-2mm}
\caption{
\textbf{Examples from the Real-annotated dataset.} We visualize both the segmentation masks as well as the boundary polygons. 
}
\label{fig:real-samples}
\vspace{-3mm}
\end{figure*}

%% file: fig/supp/sim-annotated-samples.tex
\begin{figure*}[t!]
\begin{center}
\includegraphics[width=1.0\linewidth]{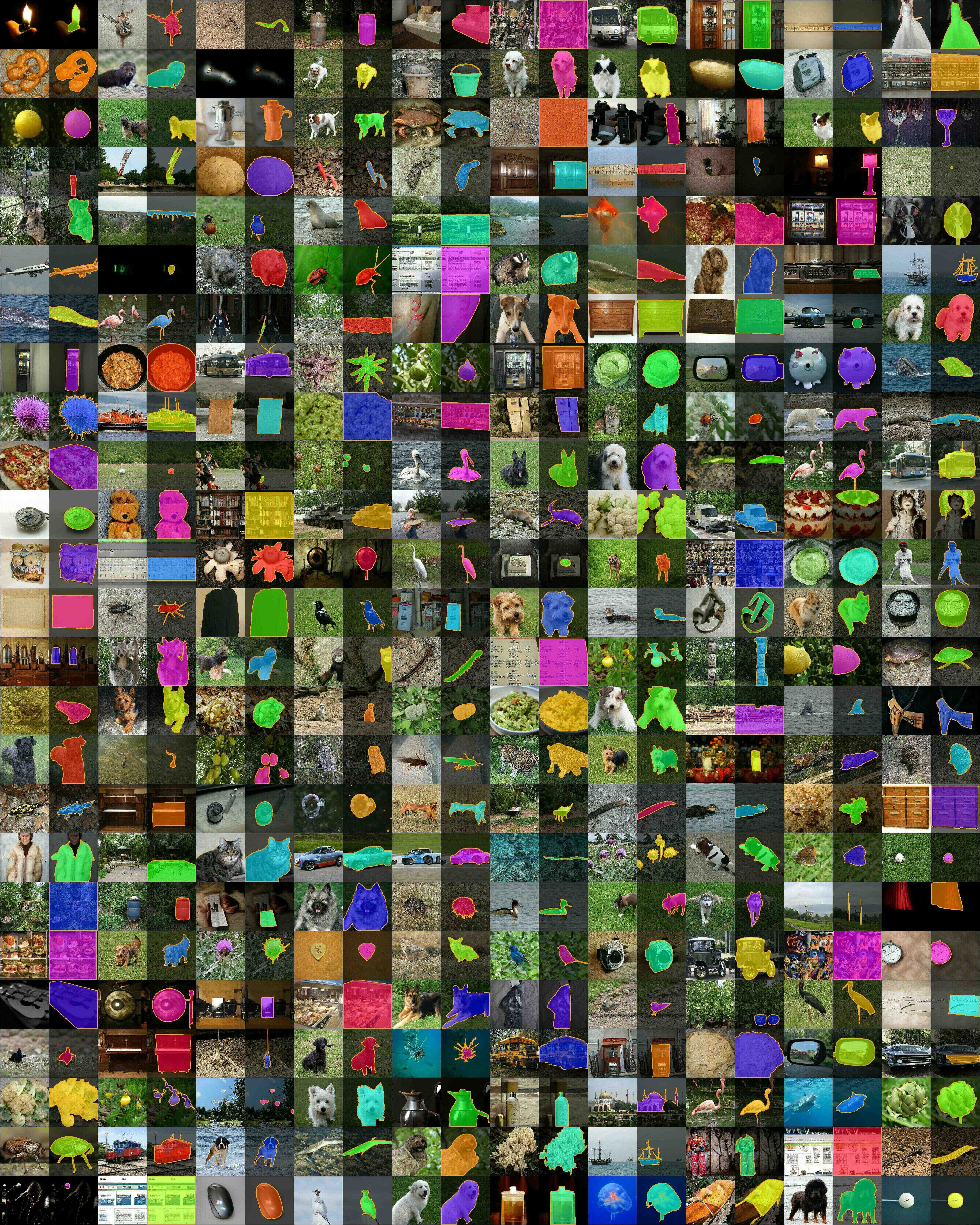}
\end{center}
\vspace{-2mm}
\caption{
\textbf{Examples from the Synthetic-annotated dataset.} We visualize both the segmentation masks as well as the boundary polygons. 
}
\label{fig:sim-samples}
\vspace{-3mm}
\end{figure*}

%% file: fig/supp/biggan-sim-random-samples.tex
\begin{figure*}[t!]
\begin{center}
\includegraphics[width=1.0\linewidth]{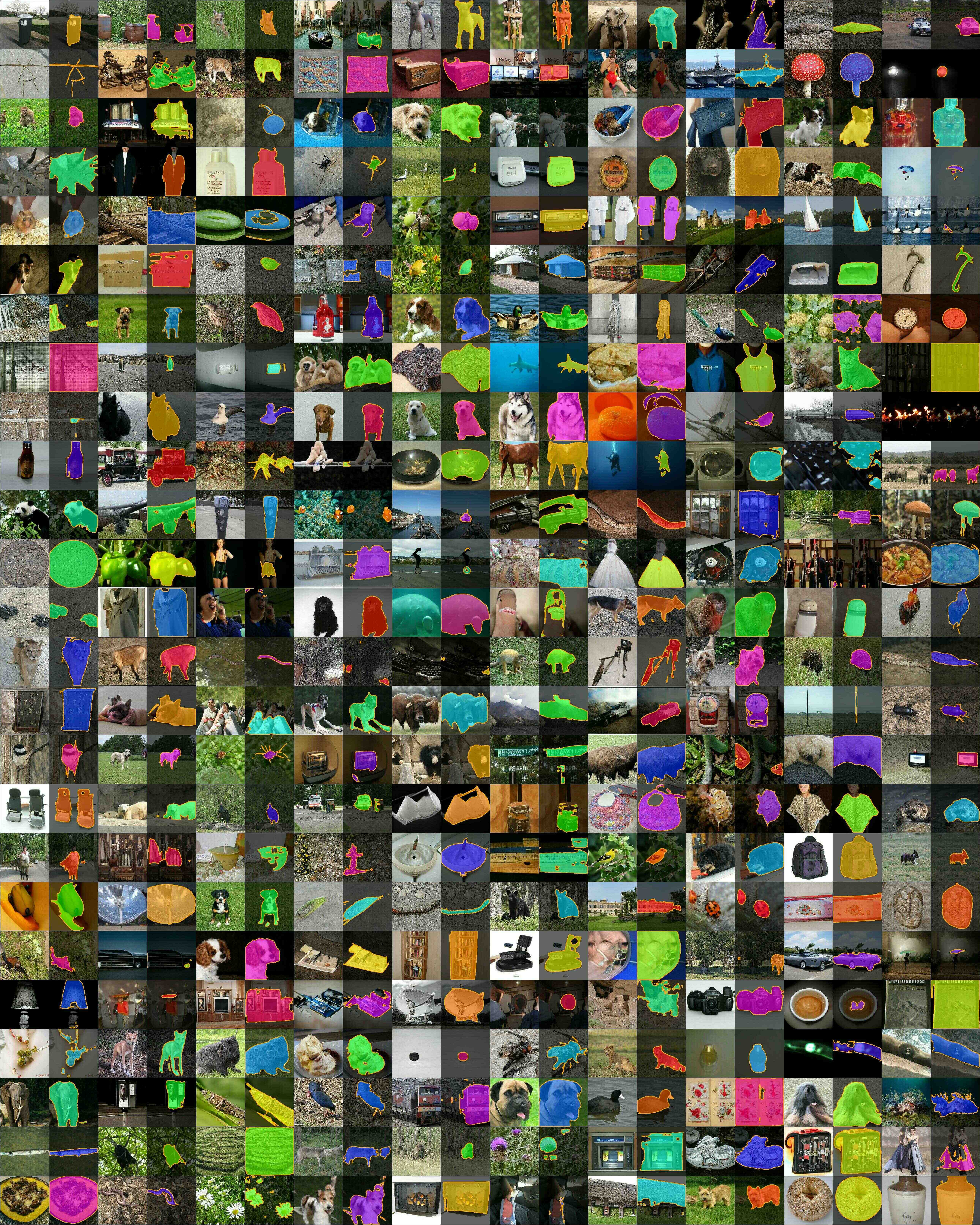}
\end{center}
\vspace{-3mm}
\caption{
\textbf{BigGAN-sim random samples.} We visualize both the segmentation masks as well as the boundary polygons. 
}
\label{fig:biggan-random-samples}
\vspace{-3mm}
\end{figure*}

%% file: fig/supp/vqgan-sim-random-samples.tex
\begin{figure*}[t!]
\begin{center}
\includegraphics[width=1.0\linewidth]{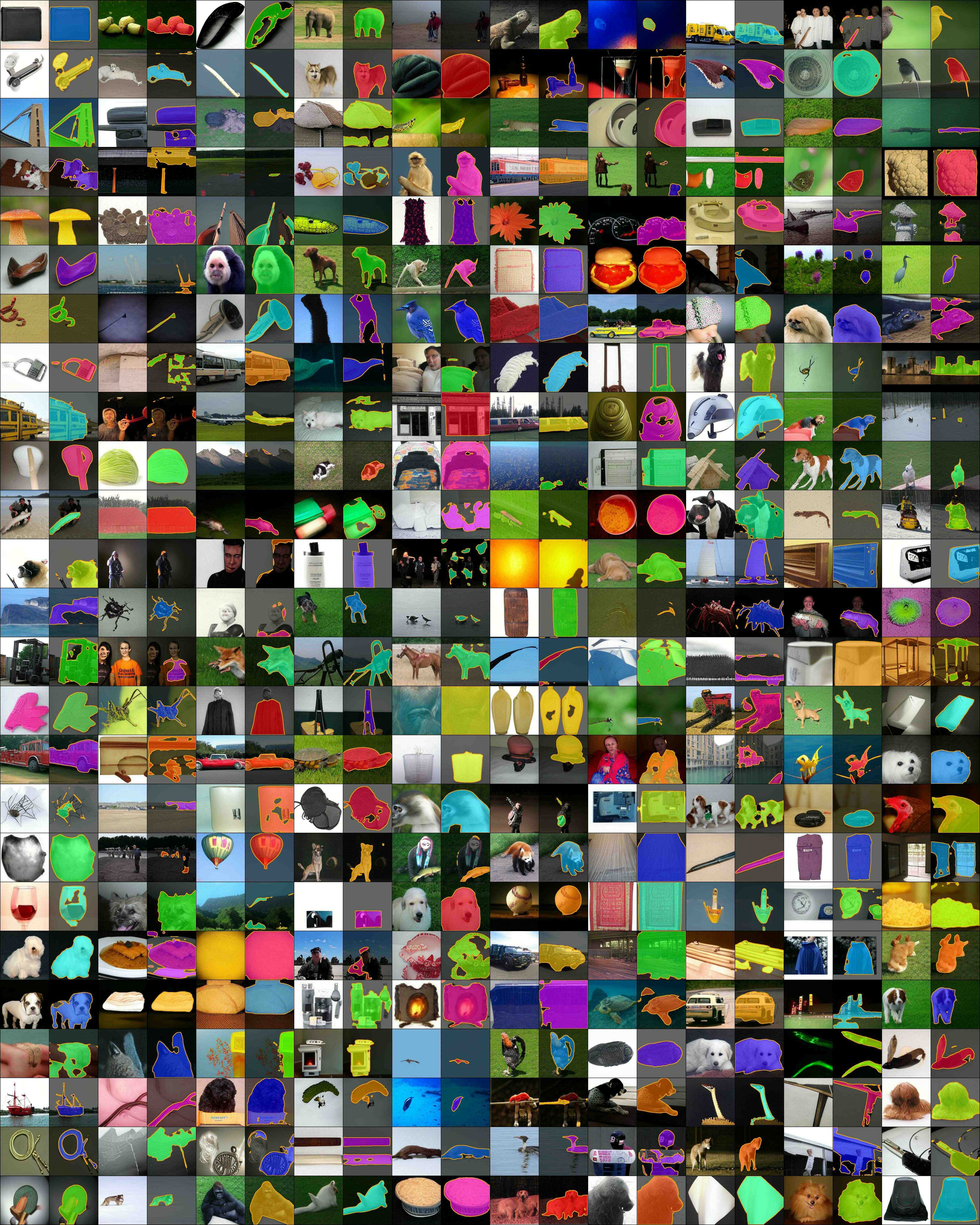}
\end{center}
\vspace{-3mm}
\caption{
\textbf{VQGAN-sim random samples.} We visualize both the segmentation masks as well as the boundary polygons. 
}
\label{fig:vqgan-random-samples}
\vspace{-3mm}
\end{figure*}

%% file: fig/supp/biggan-sim-perclass-samples.tex
\begin{figure*}[t!]
\begin{center}
\includegraphics[width=0.82\linewidth]{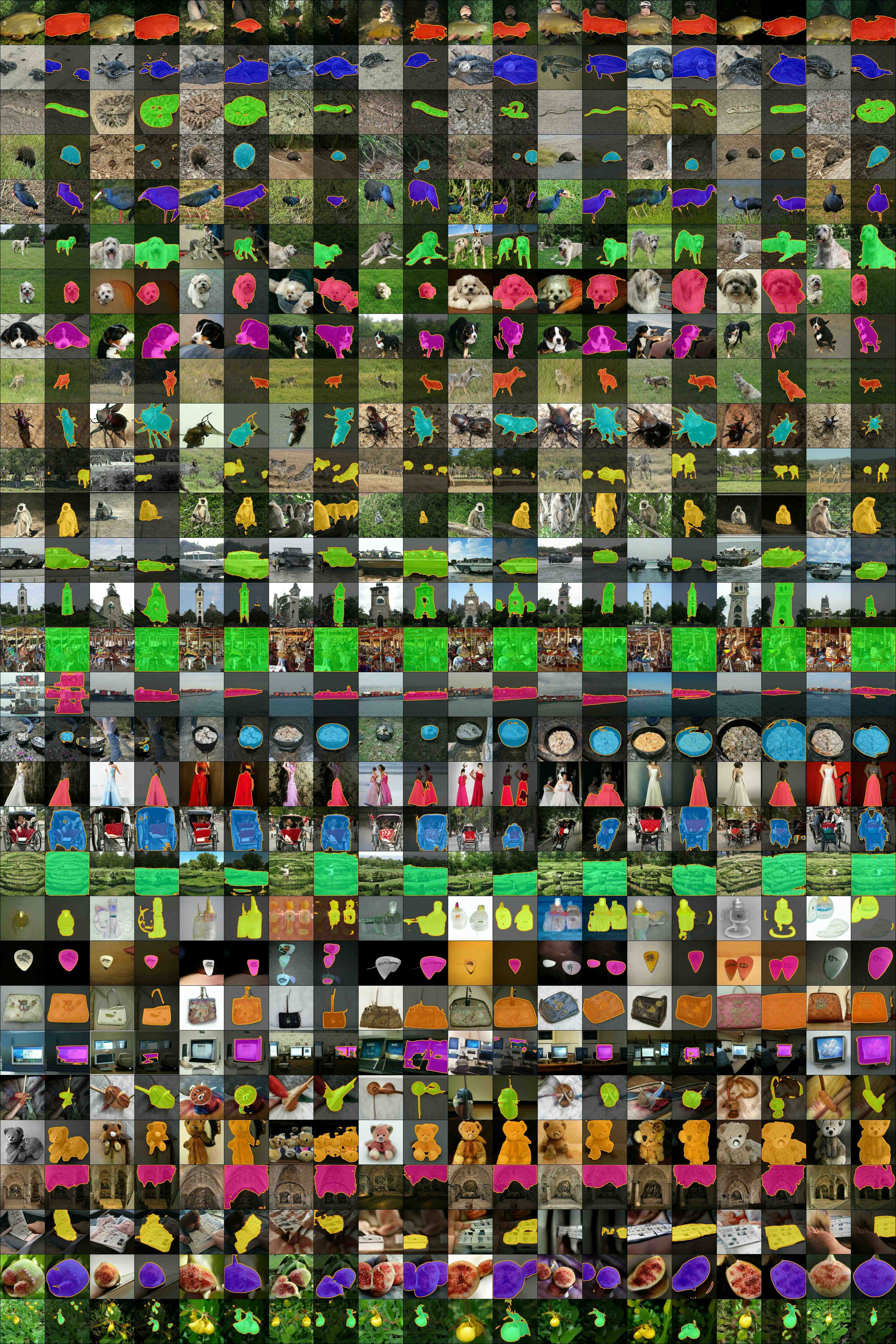}
\end{center}
\vspace{-3mm}
\caption{
\textbf{BigGAN-sim per-class samples.} We visualize both the segmentation masks as well as the boundary polygons. 
}
\label{fig:biggan-perclass-samples}
\vspace{-3mm}
\end{figure*}

%% file: fig/supp/vqgan-sim-perclass-samples.tex
\begin{figure*}[t!]
\begin{center}
\includegraphics[width=0.82\linewidth]{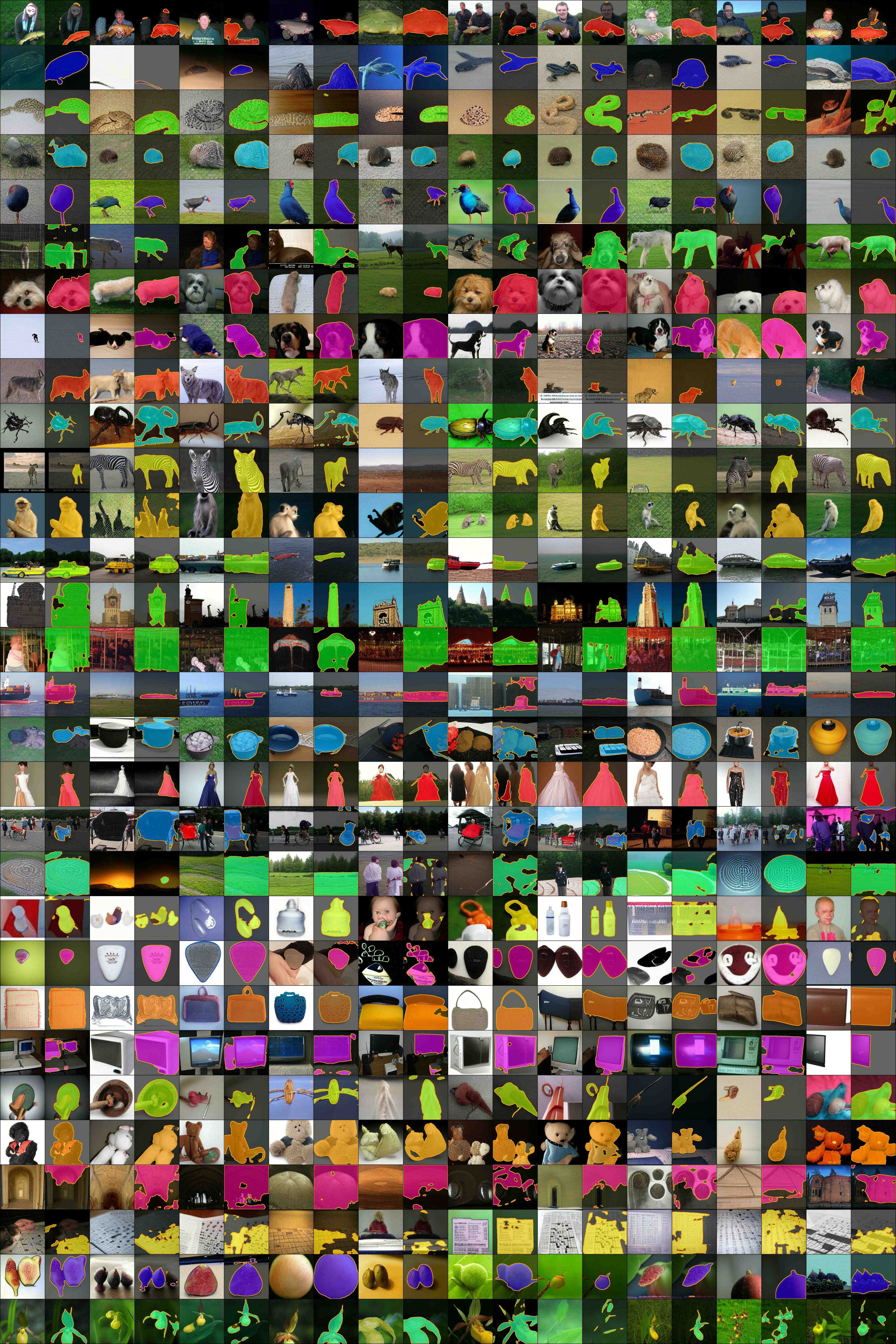}
\end{center}
\vspace{-3mm}
\caption{
\textbf{VQGAN-sim per-class samples.} We visualize both the segmentation masks as well as the boundary polygons. 
}
\label{fig:vqgan-perclass-samples}
\vspace{-3mm}
\end{figure*}

%% file: fig/supp/biggan-sim-meanshapes.tex
\begin{figure*}[t!]
\begin{center}
\includegraphics[width=1.0\linewidth]{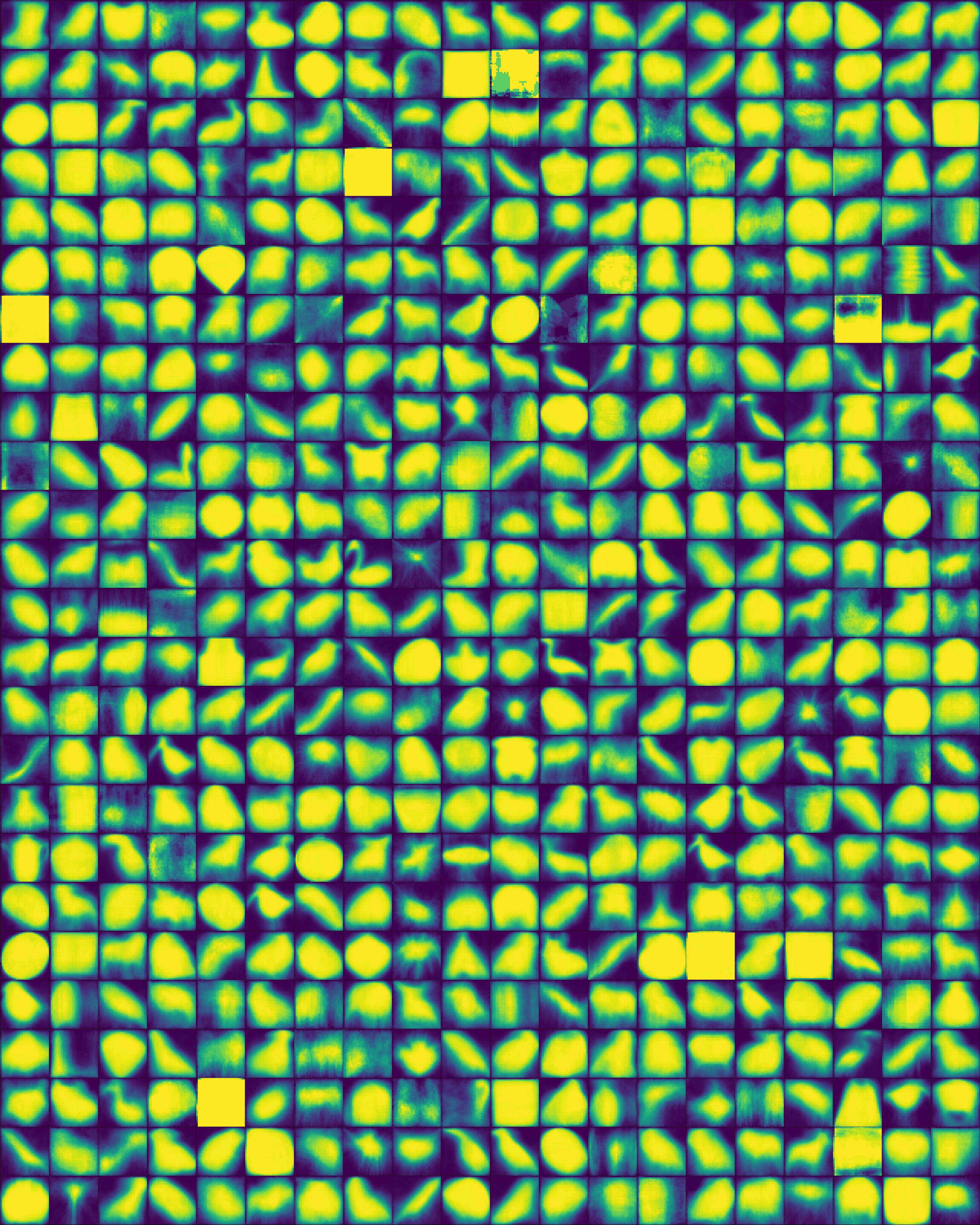}
\end{center}
\vspace{-4mm}
\caption{
\textbf{BigGAN-sim mean shapes.} Mean shapes are calculated using k-means clustering over normalized segmentation masks.
}
\label{fig:biggan-meanshapes}
\vspace{-3mm}
\end{figure*}